\documentclass[letterpaper]{article} 
\usepackage[preprint]{aaai2027} 
\usepackage[hyphens]{url} 
\usepackage{graphicx} 
\usepackage{natbib} 
\usepackage{caption} 
\usepackage{booktabs}
\usepackage{amsfonts}
\usepackage{nicefrac}
\usepackage{multirow}
\usepackage{algorithm}
\usepackage{algpseudocode}
\usepackage{needspace}
\usepackage{amsmath}
\usepackage{amssymb}

\makeatletter
\newenvironment{breakablealgorithm}
{%
  \par\addvspace{\intextsep}%
  \refstepcounter{algorithm}%
  \hrule height .8pt depth 0pt\kern 2pt%
  \renewcommand{\caption}[2][\relax]{%
    \textbf{Algorithm~\thealgorithm} ##2\par%
    \ifx\relax##1\relax%
      \addcontentsline{loa}{algorithm}{\protect\numberline{\thealgorithm}##2}%
    \else%
      \addcontentsline{loa}{algorithm}{\protect\numberline{\thealgorithm}##1}%
    \fi%
    \kern 2pt\hrule\kern 2pt%
  }%
}
{%
  \kern 2pt\hrule\par\addvspace{\intextsep}%
}
\makeatother

\title{GPrune-LLM: Generalization-Aware Structured Pruning for Large Language Models}

\author{
Xiaoyun Liu\textsuperscript{\rm 1},
Divya Saxena\textsuperscript{\rm 2},
Jiannong Cao\textsuperscript{\rm 1},
Yuqing Zhao\textsuperscript{\rm 1},
Yiying Dong\textsuperscript{\rm 1},
Penghui Ruan\textsuperscript{\rm 1}
}
\affiliations{
\textsuperscript{\rm 1}Department of Computing, The Hong Kong Polytechnic University, Hong Kong\\
\textsuperscript{\rm 2}School of Artificial Intelligence and Data Science, IIT Jodhpur, Jodhpur 342030, India\\
xiaoyun.liu@connect.polyu.hk, divyasaxena@iitj.ac.in, csjcao@comp.polyu.edu.hk,\\
csyzhao1@comp.polyu.edu.hk, yiying.dong@comp.polyu.edu.hk, penghui.ruan@connect.polyu.hk
}

\begin{document}
\maketitle
\begin{abstract}
Structured pruning is widely applied to compress large language models (LLMs), but its performance depends heavily on how neuron importance is estimated. Most existing methods rely on activation statistics from a single calibration set, which introduces calibration bias and degrades downstream cross-task generalization. We observe that neurons exhibit heterogeneous distribution sensitivity, ranging from maintaining relatively stable rankings across calibration datasets to showing substantially larger cross-dataset variation. Ignoring this heterogeneity, existing methods rank all neurons in shared spaces with a uniform scoring source, so calibration-specific neurons dominate the ranking and weakly-activated neurons are scored unreliably. To address this, we propose GPrune-LLM, a structured pruning framework that reduces calibration bias by measuring and exploiting the cross-distribution behavior of neurons for fair comparison. Specifically, we restructure the neuron ranking space into behavior-consistent local spaces, adapt the scoring source where the calibration signal is unreliable, and learn per-module sparsity allocation under a global budget. Experiments across multiple models and downstream tasks show that GPrune-LLM improves the generalization of its base pruning metrics, with gains most pronounced at high sparsity, and reduces dependence on the choice of importance metric.
\end{abstract}

\section{Introduction}
\label{sec:intro}

Large language models (LLMs) now serve as the backbone of many language and multimodal systems~\cite{touvron2023llama2,grattafiori2024llama3,liu2023llava,liu2024llava15,wang2024qwen2vl,kim2024openvla}. As model scale continues to grow, the increasing computational and memory demands have intensified the need for effective compression. Structured pruning~\cite{ma2023llmpruner,ashkboos2024slicegpt,an2024flap,ban2025gap} is especially attractive, as it removes hardware-friendly units such as neurons and attention heads while preserving dense execution on standard accelerators.

\begin{figure}[H]
    \centering
    \includegraphics[width=0.80\linewidth]{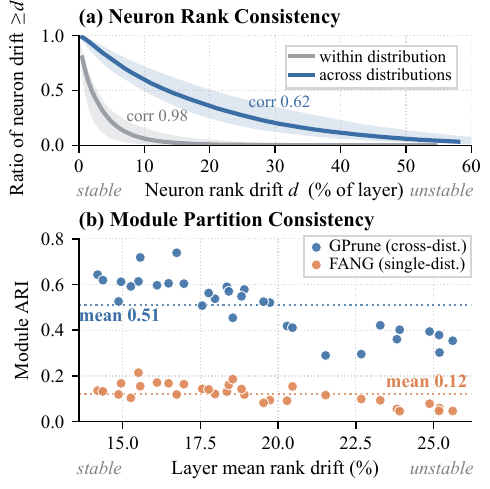}
    \caption{Neuron importance changes with the calibration distribution, and neurons differ in their sensitivity to it. \textbf{(a)} Per-neuron importance rank drift $d$ under a calibration swap, over all FFN layers. Re-sampling the same distribution (grey) barely shifts ranks (median rank correlation $0.98$), whereas switching WikiText2$\leftrightarrow$PTB (blue) causes large and heterogeneous drift ($0.62$), with some neurons drifting sharply while others remain nearly fixed. \textbf{(b)} Per-layer agreement (Adjusted Rand Index, ARI) between the neuron partitions produced before and after the same calibration swap. Modularizing neurons by cross-distribution behavior (GPrune) is more consistent than functional grouping estimated from a single distribution (FANG) (mean ARI $0.51$ vs.\ $0.12$).}
    \label{fig:fig1}
\end{figure}

However, the effectiveness of structured pruning depends critically on neuron importance estimation. Activation-based metrics have become the dominant choice because they measure parameter utilization under realistic inputs~\cite{frantar2023sparsegpt,sun2024wanda,ma2023llmpruner,an2024flap}. In practice, these metrics are computed from a small calibration set, so pruning decisions are implicitly conditioned on the calibration distribution. When that distribution does not reflect downstream usage, the induced bias can substantially degrade the generalization performance of the pruned model.

Prior studies~\cite{bandari2024c4optimal,ji2025beware} show that calibration data selection can dominate method-level differences at high sparsity. This has motivated a largely data-centric line of remedies that reduce calibration bias by improving the calibration set through careful selection~\cite{ji2025beware} or iterative augmentation~\cite{wu2026iterative}. A recent exception, FANG~\cite{yu2025fang}, instead acts at the mechanism level. It clusters neurons into functional groups by the semantic context they process and computes a fairer importance score by up-weighting the function-relevant tokens. However, this grouping and weighting are derived from a single calibration distribution, and may not reflect how a neuron's importance behaves across distributions.

We therefore directly examine how neuron importance rankings vary across calibration distributions (Fig.~\ref{fig:fig1}). We find that the cross-distribution behavior of neurons is highly heterogeneous: some retain similar importance rank across distributions, whereas others drift substantially (Fig.~\ref{fig:fig1}a). Moreover, functional modules formed from a single calibration set are only weakly consistent across distributions (Fig.~\ref{fig:fig1}b).

This heterogeneity is problematic for existing methods. When neurons of different distribution sensitivity are compared in a shared ranking space, calibration-specific neurons can dominate and suppress more broadly useful ones, causing cross-behavior ranking interference. Additionally, distribution-sensitive neurons that receive insufficient signal from the current calibration set may be ranked unreliably, leading to local ranking distortion. 

To address these issues, we propose GPrune-LLM, a structured pruning framework that reduces calibration bias by explicitly exploiting the cross-distribution behavior of neurons to restructure importance comparison into fair local ranking spaces. Specifically, we partition neurons into behavior-consistent modules by jointly considering general parameter similarity with observable cross-dataset rank drift. This ensures a fairer ranking competition among neurons, mitigating cross-behavior ranking interference. We then identify modules whose ranking is unreliable under the current calibration distribution and route each to the distribution whose local ranking is more stable, mitigating local ranking distortion. Finally, we adaptively learn per-module sparsity under a global budget, guided by cross-distribution supervision.

Our contributions are as follows:
\begin{itemize}
    \item We show that neurons exhibit heterogeneous cross-distribution behavior, which can be leveraged to reduce calibration bias in structured pruning.
    \item We propose GPrune-LLM, a structured pruning framework that reduces calibration bias by restructuring importance comparison at the module level according to neurons' cross-distribution behavior.
    \item Across multiple models and sparsity levels, GPrune-LLM improves the downstream generalization of models pruned with existing importance metrics, with larger gains at high sparsity, and is less sensitive to the choice of metric.
\end{itemize}

\section{Related Work}

\subsection{LLM Pruning}

LLM pruning methods can be broadly divided into unstructured and structured approaches. Unstructured methods remove individual weights and mainly differ in importance metrics~\cite{frantar2023sparsegpt,sun2024wanda,dong2024prunerzero,zhang2024plugandplay} or sparsity allocation strategies~\cite{yin2024owl,chen2025dlp}. However, their practical speedup depends on hardware support for sparse execution.

Structured pruning removes entire architectural units for dense-hardware acceleration, with existing methods focusing on structure identification, importance estimation, sparsity allocation, and accuracy recovery. For structure identification, LLM-Pruner~\cite{ma2023llmpruner} and Coupled-components Elimination~\cite{coupled2024naacl} discover coupled groups via dependency graphs and intermediate data flow, whereas SliceGPT~\cite{ashkboos2024slicegpt} and DISP-LLM~\cite{gao2024displlm} redefine the pruneable unit through orthogonal transformations and dimension-independent blocks. For importance estimation, LoRAPrune~\cite{loraprune2024acl} uses LoRA weight gradients, NIRVANA~\cite{ai2025nirvana} uses gradient-magnitude saliency under NTK stability, and HyWIA~\cite{liu2025toward} adaptively combines coarse- and fine-grained scoring. For sparsity allocation, SlimGPT~\cite{ling2024slimgpt} applies a layer-wise incremental ratio to limit error accumulation, and GAP~\cite{ban2025gap} jointly optimizes per-layer ratios under a global budget via dynamic programming. For accuracy recovery, SoBP~\cite{wei2024sobp} uses OBS-based reconstruction, while SlimLLM~\cite{guo2025slimllm} and Olica~\cite{he2025olica} use lightweight linear correction. FLAP~\cite{an2024flap} spans several of these axes, proposing fluctuation-based importance scoring, global score standardization for sparsity allocation, and bias compensation for output recovery.

Our work is complementary to these efforts. While they generally improve the pruning pipeline itself, we specifically focus on reducing the calibration bias for generalization-aware pruning.

\subsection{Calibration Bias in LLM Pruning}

Activation-based importance metrics~\cite{sun2024wanda,frantar2023sparsegpt,an2024flap} are widely used in LLM pruning as activations directly reflect neuron usage. However, these metrics condition rankings on a small calibration set, making importance rankings biased by calibration distribution. 

Recent works~\cite{williams2024calibration,bandari2024c4optimal,ji2025beware} show that calibration data composition can substantially affect pruning outcomes, especially at high sparsity. This has motivated a data-centric line of work that improves the calibration set itself through task-aligned data selection~\cite{ji2025beware}, multi-domain iterative selection~\cite{wu2026iterative}, output-preserving selection based on model output divergence~\cite{ai2025nirvana}, or joint search over calibration data and importance metrics~\cite{kong2025adapruner}. 

Our work takes a mechanism-centric direction, improving how calibration-conditioned scores are compared and converted into pruning decisions. A recent work, FANG~\cite{yu2025fang}, shares this goal but relies on a different signal. It groups neurons by the semantic context they serve and up-weights function-relevant tokens. This improves fairness in scoring, but since its grouping is estimated from a single distribution, it may not capture the more general behavior of neurons across distributions. In contrast, GPrune-LLM directly probes the cross-distribution behavior of neurons with an auxiliary calibration set, and forms behavior-consistent modules based on how stable each neuron's importance ranking is across distributions. FANG's token-weighted scoring is also complementary to our cross-distribution grouping.

\section{Methodology}

\begin{figure*}[t]
    \centering
    \includegraphics[width=0.86\textwidth]{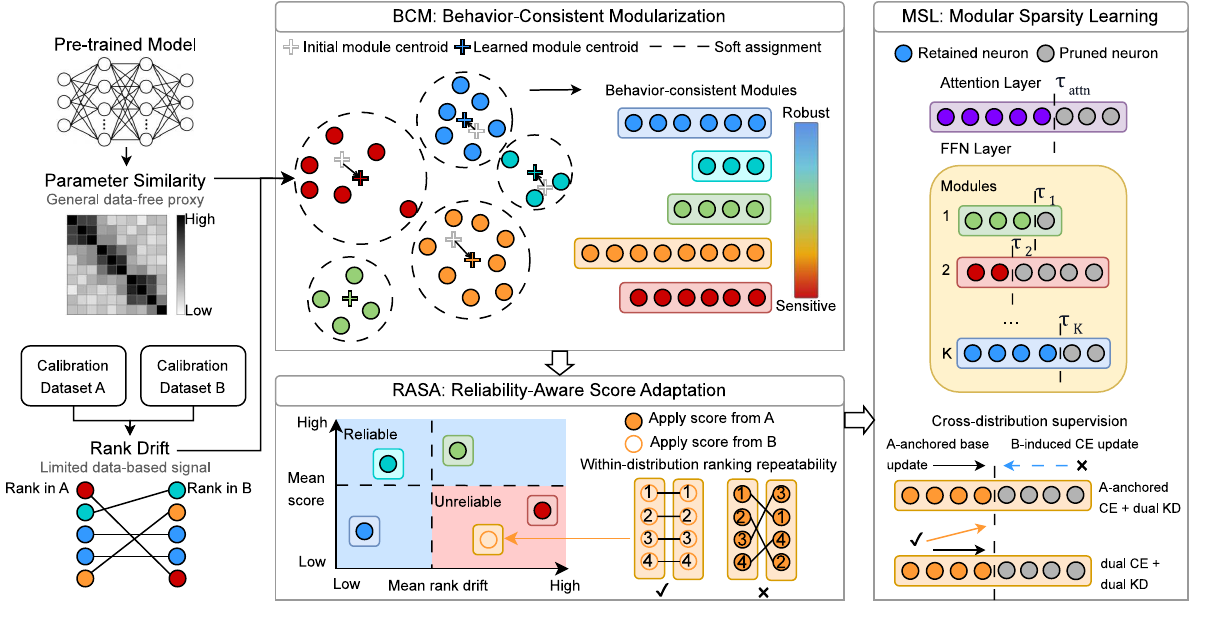}
    \caption{Overview of GPrune-LLM. Neurons are first partitioned into behavior-consistent modules by jointly modeling data-free parameter similarity and cross-distribution rank drift, with assignments refined differentiably. Each module is characterized by its mean rank drift (distribution sensitivity) and mean score (calibration-signal strength). Modules that are both distribution-sensitive and weakly supported are re-scored from the calibration distribution whose local ranking is more repeatable. Finally, module-specific sparsity is adaptively learned under a global budget with cross-distribution supervision, yielding a pruned LLM with improved generalization across diverse downstream tasks.}
    \label{fig:fig2}
\end{figure*}

In this section, we introduce GPrune-LLM, which consists of three components: Behavior-Consistent Modularization (BCM), Reliability-Aware Score Adaptation (RASA), and Modular Sparsity Learning (MSL). The overall pipeline is illustrated in Fig.~\ref{fig:fig2}. 



\subsection{Behavior-Consistent Modularization}
\label{sec:bcm}

To address cross-behavior ranking interference, we partition FFN neurons into behavior-consistent modules that serve as local ranking spaces for neurons with similar cross-distribution behavior. Cross-dataset rank drift provides a direct observable signal of this behavior. For two calibration datasets $A$ and $B$, the rank drift of neuron $i$ within its layer is:
\begin{equation}
\label{eq:rankdiff_bcdm}
d_i=\frac{|r_A(i)-r_B(i)|}{N-1},
\end{equation}
where $r_A(i)$ and $r_B(i)$ are the within-layer importance ranks computed by the chosen activation-based metric (e.g., FLAP~\cite{an2024flap}) on each dataset, and $N$ is the layer width. 

However, rank drift computed from limited calibration datasets may not fully capture intrinsic behavioral differences. Neurons with similar parameters often exhibit similar activation patterns across different inputs~\cite{zhang2022moefication}. We therefore combine rank drift with parameter similarity, a data-free proxy for co-activation affinity that depends on parameter-space structure, and regard two neurons as behavior-consistent when they are close in parameter space and share similar rank-drift profiles.

We initialize modules in two steps. We first cluster neuron parameters using KMeans~\cite{lloyd1982least} with cosine distance, selecting the module number from a candidate set $\mathcal{K}$ by the silhouette score~\cite{rousseeuw1987silhouettes}. Clusters with high internal drift heterogeneity are further split, pre-separating distribution-robust from distribution-sensitive neurons. Let $K$ denote the resulting number of modules for the current layer.

Discrete initialization cannot jointly optimize parameter similarity and rank drift consistency. We therefore refine module assignments using soft membership probabilities. For neuron $i$, the probability of belonging to module $k$ is:
\begin{equation}
\label{eq:soft_assign_bcdm}
p_{ik}=\frac{\exp\bigl(\cos(\mathbf{w}_i,\mathbf{c}_k)\bigr)}{\sum_{k'=1}^{K}\exp\bigl(\cos(\mathbf{w}_i,\mathbf{c}_{k'})\bigr)},
\end{equation}
where $\mathbf{w}_i$ is the neuron parameter and $\mathbf{c}_k$ is the module centroid, initialized from KMeans and updated during refinement. The refinement objective $\mathcal{L}_{\text{BCM}}$ is driven by three core principles: modules should remain compact in parameter space, well separated from one another, and consistent in rank-drift behavior. In practice, the compactness term is supplemented by a lightweight dispersion regularizer to prevent a few diffuse modules from dominating the refinement dynamics. 

The compactness term $\mathcal{L}_{\text{inner}}$ keeps neurons close to their assigned centroids:
\begin{equation}
\label{eq:l_inner_bcdm}
\mathcal{L}_{\text{inner}}
=\frac{1}{N}\sum_{i=1}^{N}\sum_{k=1}^{K}
p_{ik}\bigl(1-\cos(\mathbf{w}_i,\mathbf{c}_k)\bigr).
\end{equation}

The repulsion term $\mathcal{L}_{\text{rep}}$ prevents module collapse by maintaining angular separation among centroids:
\begin{equation}
\label{eq:l_rep_bcdm}
\mathcal{L}_{\text{rep}}
=\frac{1}{K(K-1)}
\sum_{k\neq k'}\cos^2(\mathbf{c}_k,\mathbf{c}_{k'}).
\end{equation}

The rank-drift consistency term $\mathcal{L}_{\text{consis}}$ groups neurons with similar drift profiles within the same module:
\begin{equation}
\label{eq:l_consis_bcdm}
\begin{aligned}
\mathcal{L}_{\text{consis}}
&=\frac{1}{K}\sum_{k=1}^{K}
\frac{\sum_{i} p_{ik}\left(d_i-\tilde{d}_k\right)^2}
{\sum_{i} p_{ik}},\\
\tilde{d}_k&=\frac{\sum_i p_{ik}d_i}{\sum_i p_{ik}},
\end{aligned}
\end{equation}
where $\tilde{d}_k$ is the soft-weighted mean drift of module $k$.

Since the rank drift magnitude alone may mix neurons that favor different distributions, we also consider the direction of the drift. This can yield more behaviorally coherent modules and sharper pruning boundaries, but can also separate neurons of similar stability and reduce within-module compactness. We augment the drift magnitude with its signed value during module refinement only when the benefit outweighs the drawback.

Collectively, these provide the structural foundation for the following modular pruning.

\subsection{Reliability-Aware Score Adaptation}
\label{sec:ramp}

To address local ranking distortion, we identify modules whose ranking is unreliable under the current calibration distribution and re-score them using the distribution that provides a more stable ranking. 

A module's ranking is most likely to be unreliable when it is both distribution-sensitive and weakly supported by the calibration signal. Therefore, we characterize each module $k$ by its mean within-module rank drift $\bar{d}^{\mathrm{m}}_{k}$ and its mean score $\bar{s}^{(A)}_{k}$ under the current calibration distribution $A$. Formally,
\begin{equation}
\label{eq:module_stats}
\begin{aligned}
\bar{d}^{\mathrm{m}}_{k}
&=
\frac{1}{|\mathcal{C}_{k}|}
\sum_{i\in\mathcal{C}_{k}}
\left|
\pi_{A,k}(i)-\pi_{B,k}(i)
\right|,\\
\bar{s}^{(A)}_{k}
&=
\frac{1}{|\mathcal{C}_{k}|}
\sum_{i\in\mathcal{C}_{k}}
s_i^{(A)},
\end{aligned}
\end{equation}
where $\pi_{D,k}(i)$ is the normalized within-module rank of neuron $i$ under calibration distribution $D$, and $s_i^{(D)}$ denotes its importance score computed from $D$.

We apply two thresholds $\delta^{\mathrm{drift}}$ and $\delta^{\mathrm{score}}$ to identify the modules to adapt, denoted as $\mathcal{K}_{\mathrm{adapt}}$. We then choose which calibration distribution to score each of them from.

Since importance scores are scaled differently across distributions, directly comparing their mean scores is insufficient to accurately identify which distribution supports a module better. We instead use a scale-free signal: within-distribution ranking repeatability. When independent subsets of the same distribution produce similar local rankings, that ordering is less sensitive to incidental sample composition and thus more trustworthy. 

Concretely, let $u_k^{(D)}$ be the expected local-rank discrepancy of module $k$ between two independent subsets drawn from distribution $D$:
\begin{equation}
\label{eq:repeatability}
u_k^{(D)}=\mathbb{E}_{D',D''\sim D}\left[\frac{1}{|\mathcal{C}_k|}\sum_{i\in\mathcal{C}_k}\left|\pi_{D',k}(i)-\pi_{D'',k}(i)\right|\right].
\end{equation}
A smaller $u_k^{(D)}$ indicates a more repeatable ranking, so for each adapted module we select the scoring source with the higher repeatability,
\begin{equation}
\label{eq:source_select}
D_k^{\star}=\arg\min_{D\in\{A,B\}} u_k^{(D)}.
\end{equation}

Let $k(i)$ denote the module of neuron $i$. The adapted importance score is
\begin{equation}
\label{eq:metric_select}
s_{i}=
\begin{cases}
s_{i}^{(D_{k(i)}^{\star})}, & k(i)\in\mathcal{K}_{\mathrm{adapt}},\\
s_{i}^{(A)}, & \text{otherwise},
\end{cases}
\end{equation}

\subsection{Modular Sparsity Learning}
\label{sec:msl}

After assigning reliable importance scores, MSL converts the modular ranking spaces into a concrete pruning mask by learning module-specific pruning thresholds under a global budget. Each module therefore adapts its sparsity to its local score landscape. Inspired by dynamic sparse training~\cite{liu2020dynamic}, we parameterize a learnable threshold $\tau_{k}$ per module. For attention heads, which are not modularized, a single threshold $\tau^{\mathrm{attn}}$ is learned per layer instead. The pruning masks for FFN neuron $i$ and attention head $h$ are:
\begin{equation}
\label{eq:mask}
\begin{aligned}
m_{i} &= S\!\left(s_{i}-\tau_{k(i)}\right),\\
m^{\mathrm{attn}}_{h} &= S\!\left(s^{\mathrm{attn}}_{h}-\tau^{\mathrm{attn}}\right),
\end{aligned}
\end{equation}
where $S(x)=\mathbf{1}[x\ge 0]$ is the binary step function, $s_i$ is the importance score assigned to FFN neuron $i$, and $s^{\mathrm{attn}}_{h}$ is the importance score of attention head $h$. A straight-through estimator (STE)~\cite{hubara2016binarized} is used for gradient propagation through the discrete masks.

The optimization objective $\mathcal{L}_{\mathrm{prune}}$ consists of a performance term $\mathcal{L}_{\mathrm{perf}}$, which combines cross-entropy (CE) with knowledge distillation (KD)~\cite{hinton2015distilling} from the unpruned model, and a sparsity-constraint term $\mathcal{L}_{\mathrm{constr}}$. To improve cross-distribution generalization, MSL applies KD on both distributions, whose soft targets rarely conflict, and incorporates the sharper CE supervision from the auxiliary distribution according to its alignment with the primary update across calibration subsets. Meanwhile, $\mathcal{L}_{\mathrm{constr}}$ enforces the global retention ratio to the target $R^\star$ throughout threshold learning.

A fixed penalty weight for the sparsity-constraint term can cause oscillation near the target and slow convergence when far from it. We therefore adopt an augmented Lagrangian formulation to balance stable constraint enforcement near the target with stronger corrective updates when the retention ratio deviates substantially. Let $g = \hat R - R^\star$ denote the deviation from the target retention ratio, where $\hat R$ is the estimated retention ratio from the STE masks. The sparsity-constraint term is:
\begin{equation}
\label{eq:l_constr}
\mathcal{L}_{\mathrm{constr}}
=\lambda_c g+\frac{\rho}{2}g^2,
\end{equation}
where $\lambda_c$ is a Lagrange multiplier updated by dual ascent and $\rho$ is the penalty coefficient.

\section{Experiments}
\subsection{Experimental Settings}

We evaluate GPrune-LLM on LLaMA2-7B and Vicuna1.5-7B. We compare against LLM-Pruner~\cite{ma2023llmpruner} without LoRA fine-tuning, Wanda-sp~\cite{sun2024wanda,an2024flap}, FLAP~\cite{an2024flap}, and FANG~\cite{yu2025fang}. GPrune-LLM is applied on top of Wanda-sp, FLAP, and FANG's token-weighted FLAP as base importance metrics, and no post-pruning fine-tuning is applied to any method. We report language modeling perplexity on WikiText2~\cite{merity2016pointer} and PTB~\cite{marcus1993building}, together with zero-shot average accuracy over seven commonsense benchmarks using the EleutherAI LM Evaluation Harness~\cite{gao2021lmharness}. Unless otherwise specified, WikiText2 is used as the primary calibration dataset and PTB is used as the auxiliary one. The additional cost of GPrune-LLM lies mainly in BCM and MSL. On a single A6000, BCM takes about 1 hour and each MSL epoch about 40 minutes, with most experiments using 2--3 epochs. Experiments on LLaMA2-13B and detailed experimental settings are provided in Appendices~\ref{sec:llama2_13b_scaling} and~\ref{sec:detailed_settings}, respectively.

\begin{table*}[!t]
\centering
\fontsize{9}{10.4}\selectfont
\setlength{\tabcolsep}{1.8pt}
\begin{tabular}{@{}c l | c c | c c c c c c c | c@{}}
\hline
Ratio & Method & WT2 PPL$\downarrow$ & PTB PPL$\downarrow$ & BoolQ & PIQA & Hella. & Wino. & ARC-e & ARC-c & OBQA & Avg. Acc.$\uparrow$ \\
\hline
100\% & LLaMA2-7B & 12.19 & 48.35 & 71.10 & 78.40 & 72.96 & 67.25 & 69.36 & 40.53 & 40.80 & 62.91 \\
\hline
80\% & LLM-Pruner & 19.23 & 72.62 & 62.26 & \underline{76.28} & 65.82 & 60.85 & 63.85 & 37.46 & 39.20 & 57.96 \\
 & Wanda-sp & 17.50 & 67.01 & 46.79 & \textbf{76.50} & \underline{68.13} & 65.43 & 65.95 & \textbf{39.76} & 38.20 & 57.25 \\
 & GPrune-LLM (Wanda-sp) & 15.44$\uparrow$ & \textbf{56.95}$\uparrow$ & 60.76$\uparrow$ & 76.22 & \textbf{68.67}$\uparrow$ & \underline{65.51}$\uparrow$ & 64.18 & 37.80 & \underline{39.80}$\uparrow$ & 58.99$\uparrow$ \\
 & FLAP & 16.41 & 61.61 & 52.45 & 73.61 & 62.56 & 62.12 & 59.13 & 34.81 & 39.20 & 54.84 \\
 & GPrune-LLM (FLAP) & 15.55$\uparrow$ & 59.56$\uparrow$ & 64.16$\uparrow$ & 74.76$\uparrow$ & 66.49$\uparrow$ & 62.19$\uparrow$ & 62.50$\uparrow$ & 36.35$\uparrow$ & 39.40$\uparrow$ & 57.98$\uparrow$ \\
 & FANG & \underline{15.43} & 61.55 & \underline{67.16} & 76.22 & 67.34 & \textbf{65.67} & \underline{66.04} & 38.14 & \textbf{40.20} & \underline{60.11} \\
 & GPrune-LLM (twFLAP) & \textbf{15.33}$\uparrow$ & \underline{58.19}$\uparrow$ & \textbf{67.89}$\uparrow$ & 75.68 & 67.90$\uparrow$ & 65.43 & \textbf{67.21}$\uparrow$ & \underline{38.23}$\uparrow$ & 39.40 & \textbf{60.25}$\uparrow$ \\
\hline
70\% & LLM-Pruner & 29.40 & 111.56 & 50.15 & 71.82 & 55.64 & 55.49 & 54.67 & 31.57 & 36.40 & 50.82 \\
 & Wanda-sp & 22.94 & 79.79 & 58.78 & 71.49 & 59.44 & 55.56 & 60.27 & \textbf{37.03} & \underline{39.40} & 54.57 \\
 & GPrune-LLM (Wanda-sp) & 21.04$\uparrow$ & 75.56$\uparrow$ & 54.80 & 72.58$\uparrow$ & \textbf{62.23}$\uparrow$ & 57.22$\uparrow$ & 60.65$\uparrow$ & 35.67 & 38.20 & 54.48 \\
 & FLAP & 20.25 & 76.02 & 53.39 & 69.48 & 55.49 & 60.30 & 54.71 & 32.59 & 38.20 & 52.02 \\
 & GPrune-LLM (FLAP) & 18.97$\uparrow$ & \underline{69.94}$\uparrow$ & \underline{63.73}$\uparrow$ & 72.03$\uparrow$ & 60.06$\uparrow$ & 57.30 & 59.85$\uparrow$ & 34.98$\uparrow$ & 38.20 & 55.17$\uparrow$ \\
 & FANG & \underline{18.86} & 78.23 & 61.65 & \textbf{73.61} & \underline{61.68} & \textbf{62.12} & \textbf{63.26} & \underline{36.77} & 38.60 & \underline{56.81} \\
 & GPrune-LLM (twFLAP) & \textbf{18.76}$\uparrow$ & \textbf{66.84}$\uparrow$ & \textbf{65.02}$\uparrow$ & \underline{73.12} & 61.49 & \underline{61.25} & \underline{61.57} & 36.26 & \textbf{39.80}$\uparrow$ & \textbf{56.93}$\uparrow$ \\
\hline
50\% & LLM-Pruner & 227.85 & 459.50 & \underline{60.61} & 58.43 & 30.56 & 49.33 & 31.19 & 27.05 & 31.40 & 41.22 \\
 & Wanda-sp & 262.32 & 505.51 & 59.82 & 54.13 & 26.54 & 49.41 & 29.00 & 23.89 & 32.80 & 39.37 \\
 & GPrune-LLM (Wanda-sp) & 60.80$\uparrow$ & 191.99$\uparrow$ & 52.72 & \underline{63.93}$\uparrow$ & 42.30$\uparrow$ & 52.49$\uparrow$ & 39.86$\uparrow$ & 28.07$\uparrow$ & \textbf{35.80}$\uparrow$ & 45.02$\uparrow$ \\
 & FLAP & 42.81 & 152.49 & 39.63 & 62.13 & 39.54 & 50.36 & 35.52 & 28.24 & \underline{35.20} & 41.52 \\
 & GPrune-LLM (FLAP) & 37.02$\uparrow$ & \textbf{125.41}$\uparrow$ & 53.27$\uparrow$ & 62.84$\uparrow$ & \textbf{44.20}$\uparrow$ & \textbf{54.78}$\uparrow$ & \textbf{44.57}$\uparrow$ & \underline{29.95}$\uparrow$ & \underline{35.20} & \underline{46.40}$\uparrow$ \\
 & FANG & \textbf{36.31} & 154.23 & \textbf{61.59} & 62.02 & 39.48 & 52.41 & 41.84 & 27.65 & 35.00 & 45.71 \\
 & GPrune-LLM (twFLAP) & \underline{36.83} & \underline{129.77}$\uparrow$ & 57.34 & \textbf{65.23}$\uparrow$ & \underline{43.72}$\uparrow$ & \underline{52.72}$\uparrow$ & \underline{43.98}$\uparrow$ & \textbf{31.74}$\uparrow$ & \textbf{35.80}$\uparrow$ & \textbf{47.22}$\uparrow$ \\
\hline
\end{tabular}
\caption{Comparison on LLaMA2-7B. For each retention ratio, the best and second-best results are marked in \textbf{bold} and \underline{underlined}, respectively. $\uparrow$ denotes improvement over the corresponding base method at the same retention ratio. For language modeling perplexity, lower is better. For classification tasks accuracy, higher is better.}
\label{tab:llama2_7b_results}
\end{table*}

\begin{table}[!t]
\centering
\begin{minipage}[t]{\linewidth}
\centering
\scriptsize
\setlength{\tabcolsep}{2.5pt}
\renewcommand{\arraystretch}{0.86}
\setlength{\aboverulesep}{0.15ex}
\setlength{\belowrulesep}{0.15ex}
\begin{tabular}{c l c c c}
\toprule
Ratio & Method & WikiText2 PPL$\downarrow$ & PTB PPL$\downarrow$ & Avg Acc$\uparrow$ \\
\midrule
100\% & Vicuna1.5-7B & 16.24 & 60.78 & 63.50 \\
\midrule
80\% & LLM-Pruner & 24.90 & 86.20 & 56.51 \\
 & Wanda-sp & 23.66 & 73.67 & 59.08 \\
 & GPrune-LLM (Wanda-sp) & 20.12$\uparrow$ & \textbf{67.35}$\uparrow$ & 58.11 \\
 & FLAP & 21.06 & 76.42 & 57.88 \\
 & GPrune-LLM (FLAP) & \underline{18.95}$\uparrow$ & \underline{68.24}$\uparrow$ & 58.57$\uparrow$ \\
 & FANG & 19.30 & 74.58 & \underline{59.13} \\
 & GPrune-LLM (twFLAP) & \textbf{18.71}$\uparrow$ & 69.15$\uparrow$ & \textbf{59.63}$\uparrow$ \\
\midrule
70\% & LLM-Pruner & 36.44 & 120.82 & 51.74 \\
 & Wanda-sp & 32.49 & 98.83 & 52.55 \\
 & GPrune-LLM (Wanda-sp) & 25.34$\uparrow$ & 86.05$\uparrow$ & 54.80$\uparrow$ \\
 & FLAP & 25.26 & 90.87 & 55.35 \\
 & GPrune-LLM (FLAP) & \textbf{22.38}$\uparrow$ & \textbf{77.41}$\uparrow$ & 56.08$\uparrow$ \\
 & FANG & \underline{22.55} & 91.04 & \underline{56.33} \\
 & GPrune-LLM (twFLAP) & 22.78 & \underline{82.90}$\uparrow$ & \textbf{56.83}$\uparrow$ \\
\midrule
50\% & LLM-Pruner & 168.91 & 450.48 & 43.96 \\
 & Wanda-sp & 270.82 & 551.68 & 39.76 \\
 & GPrune-LLM (Wanda-sp) & 71.07$\uparrow$ & 304.31$\uparrow$ & 45.01$\uparrow$ \\
 & FLAP & 50.23 & 179.31 & 43.88 \\
 & GPrune-LLM (FLAP) & \textbf{43.62}$\uparrow$ & \textbf{123.00}$\uparrow$ & \textbf{48.38}$\uparrow$ \\
 & FANG & \underline{45.31} & 183.42 & 45.40 \\
 & GPrune-LLM (twFLAP) & 51.81 & \underline{143.59}$\uparrow$ & \underline{47.84}$\uparrow$ \\
\bottomrule
\end{tabular}
\captionof{table}{Compact comparison on Vicuna1.5-7B.}
\label{tab:vicuna15_7b_results}

\end{minipage}
\end{table}

\subsection{Main Results}

\noindent\textbf{Improvement over base metrics.}
GPrune-LLM consistently improves the three base importance metrics and attains the best average accuracy at every retention ratio on both LLaMA2-7B and Vicuna1.5-7B (Tables~\ref{tab:llama2_7b_results} and~\ref{tab:vicuna15_7b_results}), indicating that it better preserves the downstream generalization of pruned models. Average accuracy increases over the base metric in 16 of the 18 settings, the only exceptions are Wanda-sp at mild sparsity, where perplexity still improves. The gains also hold when GPrune-LLM is applied on top of FANG's token-weighted metric, which indicates that our cross-distribution restructuring is complementary to the token-aware scoring. 

\noindent\textbf{Larger gains at higher sparsity.}
The gains grow as pruning becomes more aggressive, where calibration bias is most harmful. Average accuracy improvements over the base metrics are modest at 80\% retention but expand markedly at 50\%. On LLaMA2-7B, GPrune-LLM lifts Wanda-sp and FLAP by 5.65 and 4.88 points, with comparable jumps on Vicuna1.5-7B (5.25 and 4.50). At 50\% retention, GPrune-LLM also cuts Wanda-sp perplexity from 262.32 to 60.80 on LLaMA2-7B ($-76.8\%$) and from 270.82 to 71.07 on Vicuna1.5-7B. This shows that our method can mitigate severe pruning failures under extreme sparsity.

\noindent\textbf{Reduced dependence on metric choice.}
Applying GPrune-LLM makes the results less sensitive to which base metric is used. The average accuracy gap between Wanda-sp and FLAP shrinks in every setting. On LLaMA2-7B at 70\% retention, it narrows from 2.55 to 0.69 points, a 73\% reduction, with comparable reductions across the other retention ratios and on Vicuna1.5-7B. This suggests that part of the performance difference between importance metrics comes from calibration-induced ranking distortion, which GPrune-LLM reduces at the structural level.




\subsection{Visualization}
\label{sec:mechanism_visualization}

\begin{figure}[!t]
    \centering
    \includegraphics[width=0.96\columnwidth,height=0.95\columnwidth]{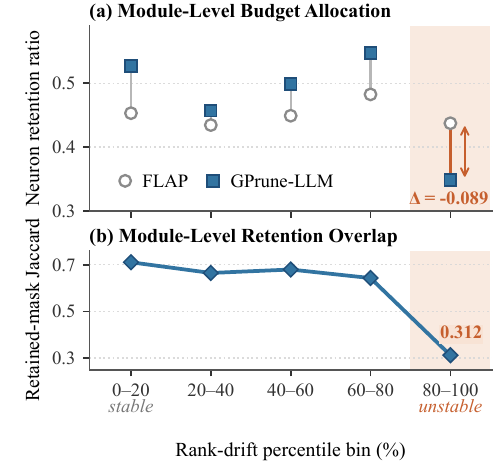}
    \caption{Visualization of how GPrune-LLM changes FLAP's pruning decisions on LLaMA2-7B at 50\% retention. \textbf{(a)} GPrune-LLM shifts the retention budget away from the most unstable modules and toward the more stable ones. \textbf{(b)} GPrune-LLM concentrates its revision of which neurons to keep on the most unstable modules.}
    \label{fig:mechanism_visualization}
\end{figure}

We visualize how GPrune-LLM changes pruning decisions relative to FLAP on LLaMA2-7B at 50\% retention, as shown in Figure~\ref{fig:mechanism_visualization}. Modules are sorted by mean cross-distribution rank drift and grouped into five equal-frequency percentile bins, from the most stable 0--20\% bin to the most unstable 80--100\% bin. 

Figure~\ref{fig:mechanism_visualization}(a) shows how GPrune-LLM reallocates the module-level retention budget. Relative to FLAP, retention increases by 0.023--0.074 in the first four bins, but decreases by 0.089 in the most unstable bin, from 0.437 to 0.349. That is, GPrune-LLM keeps more of the stable modules and prunes more from the unstable ones, as expected when cross-behavior ranking interference is mitigated.

Figure~\ref{fig:mechanism_visualization}(b) shows how GPrune-LLM changes the selection of neurons within different modules. The retained-mask Jaccard with FLAP is 0.710 in the most stable bin but drops sharply to 0.312 in the most unstable bin. This illustrates that GPrune-LLM leaves neuron selection almost unchanged in the stable modules and revises it most in the unstable ones, as expected when local ranking distortion is addressed.

\subsection{Ablation Study}

\subsubsection{Ablation of Components}

We ablate the three components of GPrune-LLM on top of FLAP at 50\% retention ratio on LLaMA2-7B. Results are shown in Table~\ref{tab:ablation_flap}. When BCM is not applied, MSL uses fixed equal-size modules, isolating the effect of learning sparsity allocation over local ranking spaces from the effect of behavior-consistent modularization.

\noindent\textbf{Modular sparsity learning.} Adding MSL alone improves WikiText2 perplexity from 42.81 to 39.56 and average accuracy from 41.52\% to 44.92\%, showing that learning local sparsity is strong even with coarse equal-size modules. However, PTB perplexity worsens from 152.49 to 159.78, indicating that this decision layer alone does not consistently improve cross-distribution perplexity when the local spaces are not behavior-consistent.

\noindent\textbf{Behavior-consistent modularization.} Adding BCM on top of MSL further improves all metrics, reducing WikiText2 perplexity from 39.56 to 37.65, PTB perplexity from 159.78 to 147.31, and raising average accuracy from 44.92\% to 45.18\%. This indicates that BCM provides a better structural basis for MSL. The gain nonetheless remains modest, as local ranking distortion within certain modules still limits the benefit of the improved structure.

\noindent\textbf{Reliability-aware score adaptation.} RASA provides an additional large gain on top of BCM and MSL, improving average accuracy to 46.40\% and reducing WikiText2 and PTB perplexity to 37.02 and 125.41, respectively. By re-scoring the modules where the current calibration signal is unreliable, RASA relieves local ranking distortion, which may let the behavior-consistent structure be more fully exploited and finally contribute to better generalization.

\begin{table}[!htbp]
\centering
\footnotesize
\setlength{\tabcolsep}{3.0pt}
\renewcommand{\arraystretch}{1.05}
\begin{tabular}{@{}c c c c c c@{}}
\toprule
BCM & RASA & MSL & WT2 PPL$\downarrow$ & PTB PPL$\downarrow$ & Avg. Acc.$\uparrow$ \\
\midrule
 &  &  & 42.81 & 152.49 & 41.52 \\
 &  & \checkmark & 39.56 & 159.78 & 44.92 \\
\checkmark &  & \checkmark & \underline{37.65} & \underline{147.31} & \underline{45.18} \\
\checkmark & \checkmark & \checkmark & \textbf{37.02} & \textbf{125.41} & \textbf{46.40} \\
\bottomrule
\end{tabular}
\caption{Component ablation of GPrune-LLM on FLAP at 50\% retention ratio on LLaMA2-7B. BCM, RASA, and MSL denote Behavior-Consistent Modularization, Reliability-Aware Score Adaptation, and Modular Sparsity Learning, respectively.}
\label{tab:ablation_flap}
\end{table}

\subsubsection{Ablation of Calibration Datasets}

We further test whether the benefit of GPrune-LLM depends on the specific calibration pair used in the main experiments. We evaluate it under all pairwise combinations of WikiText2~\cite{merity2016pointer}, PTB~\cite{marcus1993building}, and C4~\cite{raffel2020c4}, which span distinct domains. To keep the comparison aligned with the main protocol, we use C4 only as an auxiliary distribution. Results are shown in Table~\ref{tab:calib_dataset_results}.

GPrune-LLM improves over the corresponding single-dataset FLAP baseline across all reported pairings, indicating that its benefit does not depend on a specific calibration pair and transfers across different domain combinations.

The three datasets also differ in domain breadth. PTB (Wall Street Journal news) is the narrowest single-domain source, WikiText2 (curated Wikipedia articles) is broader, and C4 (heterogeneous web crawl) is the broadest. PTB and C4 thus form the most contrasting pair, and pairing them yields the largest gain in average downstream accuracy over the corresponding single-dataset baseline ($+5.52$ points). This suggests that a larger domain gap between the auxiliary and primary distributions yields more discriminative cross-distribution signals and better behavior-based module separation.

Which dataset serves as the primary distribution also matters. For the same \{WikiText2, PTB\} pair, using the more diverse WikiText2 as primary attains higher average accuracy than using PTB (46.40 vs.\ 45.83). This asymmetry is expected, since scoring and sparsity learning draw the majority of their signal from the primary distribution. A more diverse primary distribution therefore provides a stronger base.

\begin{table}[!htbp]
\centering
\scriptsize
\setlength{\tabcolsep}{2.5pt}
\renewcommand{\arraystretch}{0.86}
\setlength{\aboverulesep}{0.15ex}
\setlength{\belowrulesep}{0.15ex}
\begin{tabular}{l l c c c}
\toprule
Method & Calibration & WikiText2 PPL$\downarrow$ & PTB PPL$\downarrow$ & Avg Acc$\uparrow$ \\
\midrule
FLAP & WikiText2 & 42.81 & 152.49 & 41.52 \\
\midrule
\shortstack[l]{GPrune-LLM\\(FLAP)} & \shortstack[l]{WikiText2+\\PTB} & \textbf{37.02}$\uparrow$ & 125.41$\uparrow$ & \underline{46.40}$\uparrow$ \\
\shortstack[l]{GPrune-LLM\\(FLAP)} & \shortstack[l]{WikiText2+\\C4} & 40.93$\uparrow$ & 136.11$\uparrow$ & \textbf{46.61}$\uparrow$ \\
\midrule
FLAP & PTB & 63.87 & 119.61 & 40.55 \\
\midrule
\shortstack[l]{GPrune-LLM\\(FLAP)} & \shortstack[l]{PTB+\\WikiText2} & \underline{40.56}$\uparrow$ & \textbf{93.78}$\uparrow$ & 45.83$\uparrow$ \\
\shortstack[l]{GPrune-LLM\\(FLAP)} & \shortstack[l]{PTB+\\C4} & 55.04$\uparrow$ & \underline{117.85}$\uparrow$ & 46.07$\uparrow$ \\
\bottomrule
\end{tabular}
\caption{Comparison on different primary and auxiliary calibration datasets on FLAP at 50\% retention ratio on LLaMA2-7B.}
\label{tab:calib_dataset_results}
\end{table}

\subsection{Effect of Additional Calibration Data}

We test whether the gains of GPrune-LLM simply come from access to an extra calibration dataset. Starting from FANG with WikiText2-only scoring, we construct three multi-dataset baselines that use both WikiText2 and PTB (Table~\ref{tab:multi_dataset_effect}): averaging the importance scores computed on each dataset (mean score), averaging the per-dataset importance ranks (mean rank), or merging the two sets into a single dataset scored in one pass (pool). All three reduce perplexity on the added distribution, PTB, from 154.23 to as low as 114.20, yet they lower average downstream accuracy below the WikiText2-only baseline from 45.71 to 42.04--42.77. Therefore, naively adding a calibration dataset can fit the calibration distributions better but can not improve generalization. Using the same two datasets, GPrune-LLM instead raises average accuracy to 47.22\% while leaving WikiText2 perplexity essentially unchanged (36.83 vs.\ 36.31), indicating that the gain stems from how it exploits cross-distribution behavior to localize importance comparison rather than merely from the additional dataset.

\begin{table}[!htbp]
\centering
\scriptsize
\setlength{\tabcolsep}{2.2pt}
\renewcommand{\arraystretch}{0.86}
\setlength{\aboverulesep}{0.15ex}
\setlength{\belowrulesep}{0.15ex}
\begin{tabular}{@{}l l c c c@{}}
\toprule
Method & Scoring & WikiText2 PPL$\downarrow$ & PTB PPL$\downarrow$ & Avg Acc$\uparrow$ \\
\midrule
FANG & WT2 & \textbf{36.31} & 154.23 & \underline{45.71} \\
FANG & WT2+PTB mean score & 43.01 & \underline{115.92} & 42.18 \\
FANG & WT2+PTB mean rank & 50.29 & 144.71 & 42.77 \\
FANG & WT2+PTB pool & 42.88 & \textbf{114.20} & 42.04 \\
\midrule
\shortstack[l]{GPrune-LLM\\(twFLAP)} & WT2+PTB & \underline{36.83} & 129.77 & \textbf{47.22} \\
\bottomrule
\end{tabular}
\caption{Effect of multi-dataset scoring at 50\% retention ratio on LLaMA2-7B.}
\label{tab:multi_dataset_effect}
\end{table}

\FloatBarrier

\section{Conclusion}

We presented GPrune-LLM, a structured pruning framework that reduces calibration bias by exploiting the cross-distribution behavior of neurons for fairer local importance comparison. Motivated by the observation that neurons differ in how their importance rankings shift across calibration distributions, GPrune-LLM partitions neurons into behavior-consistent modules, re-scores the modules where the calibration signal is unreliable, and learns module-wise sparsity with cross-distribution supervision. Across multiple models and sparsity levels, GPrune-LLM consistently improves over base importance metrics, with larger gains at high sparsity and reduced dependence on the choice of metric. These results indicate that restructuring the neuron ranking space according to cross-distribution behavior is an effective way to preserve the downstream generalization of pruned LLMs.

\clearpage
\bibliography{main_arxiv_rewrite}

\clearpage
\onecolumn
\setlength{\textfloatsep}{10pt plus 2pt minus 2pt}
\setlength{\floatsep}{8pt plus 2pt minus 2pt}
\setlength{\intextsep}{8pt plus 2pt minus 2pt}
\appendix

\section{Additional Experiments and Analyses}
\label{sec:additional_mechanism_diagnostics}

\subsection{Complete Results on Vicuna1.5-7B}
\label{sec:vicuna_full_results}

Table~\ref{tab:vicuna15_7b_results_full} reports the complete per-task results on Vicuna1.5-7B, expanding the summary in the main paper.

\begin{table}[H]
\centering
{\fontsize{9}{10.4}\selectfont
\setlength{\tabcolsep}{1.8pt}
\begin{tabular}{@{}c l | c c | c c c c c c c | c@{}}
\hline
Ratio & Method & WikiText2 PPL$\downarrow$ & PTB PPL$\downarrow$ & BoolQ & PIQA & HellaSwag & WinoGrande & ARC-e & ARC-c & OBQA & Avg. Acc.$\uparrow$ \\
\hline
100\% & Vicuna1.5-7B & 16.24 & 60.78 & 75.75 & 77.80 & 71.02 & 67.64 & 69.02 & 40.87 & 42.40 & 63.50 \\
\hline
80\% & LLM-Pruner & 24.90 & 86.20 & 50.98 & \underline{75.84} & 64.14 & 59.83 & 66.12 & \underline{38.65} & 40.00 & 56.51 \\
 & Wanda-sp & 23.66 & 73.67 & \textbf{67.31} & 74.16 & 65.21 & 61.80 & 66.96 & 38.31 & 39.80 & 59.08 \\
 & GPrune-LLM (Wanda-sp) & 20.12$\uparrow$ & \textbf{67.35}$\uparrow$ & 56.76 & \textbf{75.95}$\uparrow$ & \underline{66.85}$\uparrow$ & 65.11$\uparrow$ & 65.95 & 36.95 & 39.20 & 58.11 \\
 & FLAP & 21.06 & 76.42 & \underline{66.88} & 72.47 & 63.17 & 64.48 & 63.55 & 34.81 & 39.80 & 57.88 \\
 & GPrune-LLM (FLAP) & \underline{18.95}$\uparrow$ & \underline{68.24}$\uparrow$ & 59.11 & 75.03$\uparrow$ & \textbf{67.26}$\uparrow$ & 65.27$\uparrow$ & 65.57$\uparrow$ & 36.18$\uparrow$ & \textbf{41.60}$\uparrow$ & 58.57$\uparrow$ \\
 & FANG & 19.30 & 74.58 & 60.31 & 75.41 & 66.17 & \underline{65.43} & \underline{67.38} & \textbf{38.82} & \underline{40.40} & \underline{59.13} \\
 & GPrune-LLM (twFLAP) & \textbf{18.71}$\uparrow$ & 69.15$\uparrow$ & 66.33$\uparrow$ & 74.65 & 66.22$\uparrow$ & \textbf{65.98}$\uparrow$ & \textbf{67.85}$\uparrow$ & 36.60 & 39.80 & \textbf{59.63}$\uparrow$ \\
\hline
70\% & LLM-Pruner & 36.44 & 120.82 & 43.94 & \textbf{73.45} & 56.13 & 57.30 & 58.89 & 34.30 & 38.20 & 51.74 \\
 & Wanda-sp & 32.49 & 98.83 & 44.01 & 71.33 & 57.85 & 56.51 & 62.21 & \textbf{37.12} & 38.80 & 52.55 \\
 & GPrune-LLM (Wanda-sp) & 25.34$\uparrow$ & 86.05$\uparrow$ & 51.71$\uparrow$ & 71.65$\uparrow$ & \textbf{62.64}$\uparrow$ & 61.09$\uparrow$ & 61.03 & 36.09 & 39.40$\uparrow$ & 54.80$\uparrow$ \\
 & FLAP & 25.26 & 90.87 & \textbf{66.18} & 69.86 & 57.56 & \underline{62.35} & 59.97 & 33.11 & 38.40 & 55.35 \\
 & GPrune-LLM (FLAP) & \textbf{22.38}$\uparrow$ & \textbf{77.41}$\uparrow$ & 56.39 & \underline{72.52}$\uparrow$ & 61.83$\uparrow$ & 61.80 & 63.30$\uparrow$ & 36.52$\uparrow$ & \textbf{40.20}$\uparrow$ & 56.08$\uparrow$ \\
 & FANG & \underline{22.55} & 91.04 & 53.85 & 72.09 & 61.83 & \textbf{63.69} & \textbf{65.70} & \textbf{37.12} & \underline{40.00} & \underline{56.33} \\
 & GPrune-LLM (twFLAP) & 22.78 & \underline{82.90}$\uparrow$ & \underline{63.24}$\uparrow$ & 71.33 & \underline{62.07}$\uparrow$ & 61.17 & \underline{63.97} & \underline{36.60} & 39.40 & \textbf{56.83}$\uparrow$ \\
\hline
50\% & LLM-Pruner & 168.91 & 450.48 & \textbf{62.39} & 60.23 & 34.12 & 53.35 & 35.82 & 27.82 & 34.00 & 43.96 \\
 & Wanda-sp & 270.82 & 551.68 & 60.83 & 56.64 & 27.38 & 50.12 & 29.12 & 23.04 & 31.20 & 39.76 \\
 & GPrune-LLM (Wanda-sp) & 71.07$\uparrow$ & 304.31$\uparrow$ & 44.04 & \underline{64.09}$\uparrow$ & 42.49$\uparrow$ & 52.33$\uparrow$ & 46.68$\uparrow$ & \textbf{31.66}$\uparrow$ & 33.80$\uparrow$ & 45.01$\uparrow$ \\
 & FLAP & 50.23 & 179.31 & 38.50 & 63.22 & 41.22 & 53.28 & 46.97 & 29.35 & \underline{34.60} & 43.88 \\
 & GPrune-LLM (FLAP) & \textbf{43.62}$\uparrow$ & \textbf{123.00}$\uparrow$ & 58.87$\uparrow$ & \textbf{64.80}$\uparrow$ & \textbf{44.02}$\uparrow$ & 52.80 & \textbf{50.17}$\uparrow$ & \underline{30.80}$\uparrow$ & \textbf{37.20}$\uparrow$ & \textbf{48.38}$\uparrow$ \\
 & FANG & \underline{45.31} & 183.42 & 46.94 & 62.84 & 42.23 & \underline{54.54} & \underline{48.32} & 28.33 & \underline{34.60} & 45.40 \\
 & GPrune-LLM (twFLAP) & 51.81 & \underline{143.59}$\uparrow$ & \underline{62.08}$\uparrow$ & 63.44$\uparrow$ & \underline{43.23}$\uparrow$ & \textbf{54.62}$\uparrow$ & 47.94 & 29.18$\uparrow$ & 34.40 & \underline{47.84}$\uparrow$ \\
\hline
\end{tabular}
}
\caption{Full comparison on Vicuna1.5-7B.}
\label{tab:vicuna15_7b_results_full}
\end{table}

%

\subsection{Scaling to LLaMA2-13B}
\label{sec:llama2_13b_scaling}

Beyond the 7B models, we examine whether the improvements of GPrune-LLM hold at larger scale on LLaMA2-13B (Table~\ref{tab:llama2_13b_results_full}). Since the larger model degrades little at 70--80\% retention, where the methods are hard to separate, we focus on the more challenging 30--50\% regime and omit LLM-Pruner due to its GPU memory cost.

\begin{table}[H]
\centering
 {\fontsize{9}{10.4}\selectfont
\setlength{\tabcolsep}{1.8pt}
\begin{tabular}{@{}c l | c c | c c c c c c c | c@{}}
\hline
Ratio & Method & WikiText2 PPL$\downarrow$ & PTB PPL$\downarrow$ & BoolQ & PIQA & HellaSwag & WinoGrande & ARC-e & ARC-c & OBQA & Avg. Acc.$\uparrow$ \\
\hline
100\% & LLaMA2-13B & 10.98 & 54.42 & 68.99 & 79.05 & 76.60 & 69.77 & 73.32 & 45.56 & 42.00 & 65.04 \\
\hline
50\% & Wanda-sp & 205.58 & 393.96 & 61.16 & 53.05 & 26.65 & 49.49 & 29.00 & 26.19 & 29.60 & 39.31 \\
 & GPrune-LLM (Wanda-sp) & 28.30$\uparrow$ & 126.67$\uparrow$ & 62.17$\uparrow$ & \textbf{68.17}$\uparrow$ & \textbf{54.20}$\uparrow$ & 55.09$\uparrow$ & 51.35$\uparrow$ & \textbf{33.62}$\uparrow$ & \underline{39.00}$\uparrow$ & \underline{51.94}$\uparrow$ \\
 & FLAP & 27.91 & 156.91 & \textbf{62.66} & 65.02 & 48.95 & \textbf{57.85} & 47.43 & 31.48 & 37.00 & 50.06 \\
 & GPrune-LLM (FLAP) & 26.54$\uparrow$ & \textbf{105.20}$\uparrow$ & 61.59 & 67.52$\uparrow$ & \underline{53.71}$\uparrow$ & 56.75 & \underline{52.15}$\uparrow$ & 32.85$\uparrow$ & 38.20$\uparrow$ & 51.82$\uparrow$ \\
 & FANG & \underline{26.43} & 155.46 & \underline{62.35} & 64.85 & 50.53 & 55.64 & 50.55 & 31.23 & \textbf{39.20} & 50.62 \\
 & GPrune-LLM (twFLAP) & \textbf{25.00}$\uparrow$ & \underline{120.01}$\uparrow$ & 62.26 & \underline{67.85}$\uparrow$ & 51.22$\uparrow$ & \underline{56.99}$\uparrow$ & \textbf{54.21}$\uparrow$ & \underline{33.19}$\uparrow$ & 38.40 & \textbf{52.02}$\uparrow$ \\
\hline
40\% & Wanda-sp & 249.44 & 565.56 & 61.68 & 53.48 & 26.48 & 49.64 & 27.74 & 24.91 & 30.60 & 39.22 \\
 & GPrune-LLM (Wanda-sp) & 83.47$\uparrow$ & 357.04$\uparrow$ & 61.74$\uparrow$ & 60.61$\uparrow$ & 40.44$\uparrow$ & 51.54$\uparrow$ & \underline{40.45}$\uparrow$ & 26.62$\uparrow$ & \underline{35.80}$\uparrow$ & \underline{45.31}$\uparrow$ \\
 & FLAP & 45.29 & 263.68 & \underline{62.08} & 52.98 & 36.90 & 52.25 & 38.05 & 27.39 & 34.80 & 43.49 \\
 & GPrune-LLM (FLAP) & \underline{45.21}$\uparrow$ & \textbf{154.85}$\uparrow$ & 49.97 & \textbf{61.86}$\uparrow$ & \underline{40.83}$\uparrow$ & 52.01 & 38.72$\uparrow$ & \underline{29.61}$\uparrow$ & 35.40$\uparrow$ & 44.06$\uparrow$ \\
 & FANG & \textbf{39.51} & 230.79 & 57.98 & 59.90 & 39.89 & \underline{52.49} & 40.40 & 27.73 & 35.00 & 44.77 \\
 & GPrune-LLM (twFLAP) & 46.04 & \underline{172.63}$\uparrow$ & \textbf{62.11}$\uparrow$ & \underline{61.26}$\uparrow$ & \textbf{41.27}$\uparrow$ & \textbf{53.67}$\uparrow$ & \textbf{46.80}$\uparrow$ & \textbf{30.63}$\uparrow$ & \textbf{36.80}$\uparrow$ & \textbf{47.51}$\uparrow$ \\
\hline
30\% & Wanda-sp & 1943.76 & 3163.66 & 51.16 & 52.67 & 26.09 & 47.51 & 26.60 & \underline{27.39} & 29.20 & 37.23 \\
 & GPrune-LLM (Wanda-sp) & 362.63$\uparrow$ & 1023.02$\uparrow$ & 52.45$\uparrow$ & \underline{56.58}$\uparrow$ & \underline{31.20}$\uparrow$ & \underline{52.01}$\uparrow$ & 34.01$\uparrow$ & 26.88 & \textbf{34.80}$\uparrow$ & \underline{41.13}$\uparrow$ \\
 & FLAP & 2676.82 & 3872.98 & 37.74 & 53.23 & 25.95 & 48.78 & 27.48 & \textbf{28.92} & 29.60 & 35.96 \\
 & GPrune-LLM (FLAP) & 119.35$\uparrow$ & \textbf{339.25}$\uparrow$ & 47.68$\uparrow$ & 56.26$\uparrow$ & 30.80$\uparrow$ & \textbf{53.43}$\uparrow$ & \textbf{37.21}$\uparrow$ & 27.22 & 32.60$\uparrow$ & 40.74$\uparrow$ \\
 & FANG & \textbf{75.39} & 392.39 & \underline{54.53} & \underline{56.42} & 30.72 & 50.36 & 33.71 & 25.60 & \underline{32.00} & 40.48 \\
 & GPrune-LLM (twFLAP) & \underline{90.77} & \underline{391.91}$\uparrow$ & \textbf{59.51}$\uparrow$ & \textbf{57.29}$\uparrow$ & \textbf{31.58}$\uparrow$ & 51.85$\uparrow$ & \underline{34.05}$\uparrow$ & 26.62$\uparrow$ & \underline{34.20}$\uparrow$ & \textbf{42.16}$\uparrow$ \\
\hline
\end{tabular}
}
\caption{Full comparison on LLaMA2-13B. WikiText2 is abbreviated as WT2 elsewhere in the supplement. Throughout the supplement, \textbf{bold} and \underline{underlined} values denote the best and second-best entries, respectively, and $\uparrow$ indicates improvement over the corresponding base method. Lower perplexity and higher accuracy are better.}
\label{tab:llama2_13b_results_full}
\end{table}

The improvements of GPrune-LLM carry over to the 13B model. Across the three retention ratios and three base metrics, it improves average accuracy in all nine comparisons and lowers PTB perplexity in every setting, while reducing WikiText2 perplexity in seven of the nine. Even when the base method collapses, GPrune-LLM still recovers a usable model. At 50\% retention, it raises Wanda-sp accuracy from 39.31 to 51.94 ($+12.63$), and at 30\% retention, it recovers FLAP's perplexity from 2676.82/3872.98 to 119.35/339.25 and its accuracy from 35.96 to 40.74. The only two WikiText2 regressions occur with twFLAP at 40\% and 30\% retention, where PTB perplexity and average accuracy still improve, consistent with GPrune-LLM optimizing for cross-distribution generalization rather than fitting the primary calibration distribution.

\subsection{Fine-Grained Ablations of BCM, RASA, and MSL}
\label{sec:detailed_component_ablations}

Having verified the overall effectiveness across model scales, we now extend the cumulative component ablation in the main paper to the internal design choices within BCM, RASA, and MSL. Table~\ref{tab:fine_grained_ablations} separates these on LLaMA2-7B with FLAP at 50\% retention ratio: BCM modularization, RASA module and scoring-source selection, and MSL supervision.

\begin{table*}[!htbp]
\centering
{\fontsize{9}{10.4}\selectfont
\setlength{\tabcolsep}{4.2pt}
\renewcommand{\arraystretch}{1.02}
\begin{tabular}{@{}l l l c c c@{}}
\toprule
Component & Design dimension & Variant & WT2 PPL$\downarrow$ & PTB PPL$\downarrow$ & Avg. Acc.$\uparrow$ \\
\midrule
\multirow{3}{*}{BCM}
 & \multirow{3}{*}{Modularization} & Random & 38.80 & 134.15 & \underline{45.46} \\
 & & Initialization & \textbf{35.68} & \underline{133.53} & 44.00 \\
 & & Initialization + refinement & \underline{37.02} & \textbf{125.41} & \textbf{46.40} \\
\midrule
\multirow{6}{*}{RASA}
 & \multirow{4}{*}{Module selection} & None & 37.65 & 147.31 & 45.18 \\
 & & Random & 36.59 & 136.71 & 45.00 \\
 & & Rank drift & \textbf{35.44} & \textbf{124.29} & \underline{46.10} \\
 & & Rank drift + score & \underline{37.02} & \underline{125.41} & \textbf{46.40} \\
\cmidrule(lr){2-6}
 & \multirow{2}{*}{Scoring source selection} & Random & \underline{37.50} & \underline{126.25} & \underline{44.21} \\
 & & Repeatability & \textbf{37.02} & \textbf{125.41} & \textbf{46.40} \\
\midrule
\multirow{3}{*}{MSL}
 & \multirow{3}{*}{Supervision} & Primary only & \textbf{34.83} & 136.75 & \underline{45.26} \\
 & & Dual & \underline{35.16} & \underline{127.14} & 44.99 \\
 & & Adaptive & 37.02 & \textbf{125.41} & \textbf{46.40} \\
\bottomrule
\end{tabular}
}
\caption{Fine-grained component ablations on LLaMA2-7B with FLAP at 50\% retention ratio.}
\label{tab:fine_grained_ablations}
\end{table*}

\noindent\textbf{BCM.}
Here only the modularization varies, with RASA and MSL kept active. The Random baseline splits neurons into fixed equal-size modules, as in the main ablation. Initialization forms a discrete partition by clustering neuron parameters and splitting high-drift clusters. Although it lowers both perplexities, average accuracy drops to 44.00, indicating that separately imposing parameter similarity and rank-drift structure through discrete operations is insufficient to form behavior-consistent modules. Differentiable refinement jointly reconciles the two signals, increasing average accuracy to 46.40 and further reducing PTB perplexity to 125.41, with a modest increase in WikiText2 perplexity.

\noindent\textbf{RASA.}
With BCM and MSL fixed, random module selection slightly reduces average accuracy relative to no adaptation (45.00 vs.\ 45.18), showing that indiscriminately introducing the auxiliary distribution through arbitrary modules does not help. The key is to identify modules whose current rankings are actually unreliable. Rank drift locates distribution-sensitive modules, raising average accuracy to 46.10 and achieving the lowest perplexities. However, distribution sensitivity alone does not necessarily imply that the primary calibration signal is unreliable. Adding the primary score criterion excludes modules that remain well supported by the primary calibration signal. This increases average accuracy from 46.10 to 46.40 while retaining the perplexity improvements over no adaptation. The downstream gain suggests that the primary signal in these excluded modules retains some transferable information despite their distribution sensitivity. Once the modules to adapt are identified, RASA must determine which distribution provides the more reliable local ordering. Random source selection reduces accuracy to 44.21, whereas repeatability-based selection achieves 46.40.

\noindent\textbf{MSL.}
With BCM and RASA fixed, primary-only supervision achieves the lowest WikiText2 perplexity but yields weaker PTB perplexity and downstream accuracy, consistent with its emphasis on the primary distribution. Fully dual supervision improves PTB perplexity from 136.75 to 127.14 but reduces average accuracy from 45.26 to 44.99, suggesting that auxiliary supervision can introduce conflicting FFN threshold updates. By considering its alignment with the primary objective and adaptively incorporating it, MSL achieves the highest average accuracy (46.40).

\subsection{Multi-Dataset Calibration across Base Metrics}
\label{sec:multi_dataset_calib}

We next test whether the gains simply come from access to a second calibration dataset, extending the main paper's token-weighted FLAP comparison to all three base metrics. On LLaMA2-7B at 50\% retention (Table~\ref{tab:multi_dataset_calib}), we evaluate three multi-dataset baselines for Wanda-sp, FLAP, and FANG that combine WikiText2 and PTB: averaging importance scores across the two datasets (score avg.), averaging their per-dataset ranks (rank avg.), and pooling the two sets before scoring (pool).

\begin{table}[H]
\centering
{\fontsize{9}{10.4}\selectfont
\setlength{\tabcolsep}{2.6pt}
\begin{tabular}{l l c c c}
\toprule
Method & Scoring & WT2 PPL$\downarrow$ & PTB PPL$\downarrow$ & Avg. Acc.$\uparrow$ \\
\midrule
Wanda-sp              & WT2                 & 262.32 & 505.51 & 39.37 \\
Wanda-sp              & Score avg.          & 111.15 & 309.30 & 40.37 \\
Wanda-sp              & Rank avg.           & 70.30  & 198.71 & 42.01 \\
Wanda-sp              & Pool                & 275.81 & 478.19 & 39.03 \\
\shortstack[l]{GPrune-LLM\\(Wanda-sp)} & WT2+PTB             & 60.80  & 191.99 & 45.02 \\
\midrule
FLAP                  & WT2                 & 42.81 & 152.49 & 41.52 \\
FLAP                  & Score avg.          & 45.41  & 117.21 & 40.79 \\
FLAP                  & Rank avg.           & 46.27  & 119.63 & 40.98 \\
FLAP                  & Pool                & 46.57  & 120.69 & 40.68 \\
\shortstack[l]{GPrune-LLM\\(FLAP)} & WT2+PTB             & 37.02 & 125.41 & \underline{46.40} \\
\midrule
FANG                  & WT2                 & \textbf{36.31} & 154.23 & 45.71 \\
FANG                  & Score avg.          & 43.01 & \underline{115.92} & 42.18 \\
FANG                  & Rank avg.           & 50.29 & 144.71 & 42.77 \\
FANG                  & Pool                & 42.88 & \textbf{114.20} & 42.04 \\
\shortstack[l]{GPrune-LLM\\(twFLAP)} & WT2+PTB             & \underline{36.83} & 129.77 & \textbf{47.22} \\
\bottomrule
\end{tabular}
}
\caption{Comparison of multi-dataset calibration at 50\% retention ratio on LLaMA2-7B. Score avg. averages importance scores across WikiText2 and PTB. Rank avg. averages per-dataset ranks. Pool merges both calibration sets before scoring.}
\label{tab:multi_dataset_calib}
\end{table}

The same conclusion extends to all three base metrics: GPrune-LLM attains the highest average accuracy in every block (45.02, 46.40, and 47.22), and for FLAP and FANG the averaging and pooling baselines even fall below single-dataset WikiText2 scoring, confirming that naively combining two calibration datasets does not account for the gains.

\subsection{Module Agreement under Calibration-Pair Substitution}
\label{sec:cross_dataset_module_overlap}

Beyond pruning performance, we examine the module structure that BCM produces and whether it transfers across different calibration datasets. Figure~1(b) of the main paper shows that cross-distribution modularization with token-weighted FLAP is more consistent across a calibration swap than single-distribution grouping. We extend this analysis to all single-dataset substitutions among WikiText2, PTB, and C4 with FLAP. Replacing one dataset changes the rank drift signal used by BCM, so agreement between the resulting partitions measures the shared cross-distribution structure recovered across calibration pairs.

We measure agreement separately in each FFN layer. Let $\{\mathcal{C}_{\ell,k}\}_{k=1}^{K_\ell}$ and $\{\mathcal{C}'_{\ell,j}\}_{j=1}^{K'_\ell}$ denote the partitions of layer $\ell$ obtained from two calibration pairs. We report the Adjusted Rand Index (ARI), which is invariant to module labels and corrects agreement for chance. For a more intuitive assignment-level measure, let $\Pi_\ell$ denote the optimal one-to-one matching between the two sets of modules. The matched overlap in layer $\ell$ is
\begin{equation}
\mathrm{Overlap}^{\mathrm{match}}_{\ell}
=
\frac{1}{N_\ell}
\sum_{(k,j)\in\Pi_\ell}
|\mathcal{C}_{\ell,k}\cap\mathcal{C}'_{\ell,j}|,
\end{equation}
where $N_\ell$ is the number of neurons in the layer. We report the mean and standard deviation of both metrics across layers.

\begin{table}[!htbp]
\centering
{\fontsize{9}{10.4}\selectfont
\setlength{\tabcolsep}{6pt}
\begin{tabular}{lcc}
\toprule
\textbf{Dataset-pair comparison} & \textbf{ARI}$\uparrow$ & \textbf{Matched overlap}$\uparrow$ \\
\midrule
WT2--PTB vs.\ WT2--C4 & $0.575\pm0.121$ & $0.707\pm0.122$ \\
WT2--PTB vs.\ PTB--C4 & $0.582\pm0.122$ & $0.706\pm0.124$ \\
WT2--C4 vs.\ PTB--C4 & $0.583\pm0.124$ & $0.714\pm0.123$ \\
\bottomrule
\end{tabular}
}
\caption{Per-layer module agreement under calibration-pair substitution on LLaMA2-7B with FLAP. Values are mean $\pm$ standard deviation across 32 FFN layers.}
\label{tab:cross_dataset_module_overlap}
\end{table}

As shown in Table~\ref{tab:cross_dataset_module_overlap}, mean ARI ranges from 0.575 to 0.583 and matched overlap from 0.706 to 0.714 across the three substitutions. The comparable agreement across calibration-pair substitutions suggests that BCM recovers modular structure shared across calibration distributions, with about 71\% of neuron assignments preserved after aligning module labels. Together with Figure~1(b) of the main paper, these results show that cross-distribution behavior supports generalizable module organization across calibration-pair choices.

\subsection{Effect of Calibration Sample Number}
\label{sec:calib_ablation}

After varying which calibration datasets are used, we now vary how many samples we draw from each. We study how the number of calibration samples $|\mathcal{D}|$ affects GPrune-LLM, where $|\mathcal{D}|$ denotes the number of samples per dataset. Results on LLaMA2-7B with FLAP at 50\% retention ratio are reported in Table~\ref{tab:calib_ablation}.

\begin{table}[!htbp]
\centering
{\fontsize{9}{10.4}\selectfont
\setlength{\tabcolsep}{2.5pt}
\begin{tabular}{c c c c c c}
\toprule
Retention & $|\mathcal{D}|$ & Epoch & WT2 PPL$\downarrow$ & PTB PPL$\downarrow$ & Avg. Acc.$\uparrow$ \\
\midrule
\multirow{5}{*}{50\%}
 & 128  & 4 & \underline{39.23} & 132.98 & 46.36 \\
 & 256  & 7 & 39.56 & 145.52 & 45.86 \\
 & 512  & 7 & 40.17 & 147.29 & \textbf{46.71} \\
 & 1024 & 8 & 39.67 & \underline{129.07} & 45.70 \\
 & 2048 & 2 & \textbf{37.02} & \textbf{125.41} & \underline{46.40} \\
\bottomrule
\end{tabular}
}
\caption{Effect of calibration sample number on GPrune-LLM (FLAP) at 50\% retention ratio on LLaMA2-7B.}
\label{tab:calib_ablation}
\end{table}

We determine the epoch for each calibration size from its observed convergence trajectory. Once trained for sufficient epochs, the 128--1024-sample settings reach similar performance, indicating that additional optimization can partly compensate for limited calibration data. However, they require 4--8 epochs, while 2048 samples reach a stronger perplexity--accuracy trade-off in only two epochs. The intermediate sizes take the most epochs possibly because they carry more sample variation than the smaller sets but still lack the stable estimates of the largest one, so their thresholds are slower to settle. Thus, 2048 samples remain preferable for efficient and stable calibration, although smaller calibration budgets remain viable when additional optimization is affordable.

\subsection{Repeatability across Random Seeds}
\label{sec:multiseed_repeatability}

We further assess the repeatability of GPrune-LLM's improvement across random seeds, using the token-weighted FLAP metric to compare against the strongest baseline, FANG. We repeat the full pruning pipeline of both methods under three random seeds ($0,1,2$) on LLaMA2-7B at 50\% retention.

\begin{table}[!htbp]
\centering
{\fontsize{9}{10.4}\selectfont
\setlength{\tabcolsep}{5pt}
\begin{tabular}{l c c c c}
\toprule
Method & Seed & WT2 PPL$\downarrow$ & PTB PPL$\downarrow$ & Avg. Acc.$\uparrow$ \\
\midrule
\multirow{4}{*}{FANG}
 & 0 & 36.31 & 154.23 & 45.71 \\
 & 1 & 36.85 & 155.82 & 44.88 \\
 & 2 & 37.55 & 158.75 & 45.50 \\
\cmidrule(lr){2-5}
 & Mean $\pm$ Std. & \textbf{$36.90\pm0.51$} & $156.27\pm1.87$ & $45.36\pm0.35$ \\
\midrule
\multirow{4}{*}{GPrune-LLM (twFLAP)}
 & 0 & 36.83 & 129.77 & 47.22 \\
 & 1 & 36.16 & 146.90 & 45.65 \\
 & 2 & 38.45 & 144.87 & 46.60 \\
\cmidrule(lr){2-5}
 & Mean $\pm$ Std. & $37.15\pm0.96$ & \textbf{$140.51\pm7.64$} & \textbf{$46.49\pm0.64$} \\
\bottomrule
\end{tabular}
}
\caption{Seed repeatability on LLaMA2-7B at 50\% retention. Mean and standard deviation are computed over seeds 0--2.}
\label{tab:multiseed_repeatability}
\end{table}

Across all three seeds, GPrune-LLM improves average accuracy over FANG, raising the mean by 1.13 points, and lowers PTB perplexity while keeping WikiText2 perplexity comparable. Its generalization gain is therefore consistent across random seeds.

\FloatBarrier

\subsection{Practical Inference Speedup}
\label{sec:speedup}

Finally, we demonstrate the practical inference speedup of GPrune-LLM by measuring its inference throughput at 50\% retention ratio on LLaMA2-7B (Table~\ref{tab:actual_speedup}). We use FP16 inference with KV-cached greedy decoding, one warm-up run, prompt length 2048, generation length 512, batch size 8, and 10 timed repetitions. 

\begin{table}[H]
\centering
{\fontsize{9}{10.4}\selectfont
\setlength{\tabcolsep}{6pt}
\begin{tabular}{c l c c}
\toprule
\shortstack[c]{Retention\\Ratio} & Method & Params & \shortstack[c]{Tokens/s$\uparrow$} \\
\midrule
100\% & LLaMA2-7B & 6.74B & 97.04 \\
\midrule
50\% & LLM-Pruner & 3.45B & 161.69 \\
 & Wanda-sp & 3.5B & 167.44 \\
 & GPrune-LLM (Wanda-sp) & 3.49B & 160.38 \\
 & FLAP & 3.5B & 148.52 \\
 & GPrune-LLM (FLAP) & 3.49B & 154.59 \\
 & FANG & 3.53B & 159.57 \\
 & GPrune-LLM (twFLAP) & 3.51B & 153.51 \\
\bottomrule
\end{tabular}
}
\caption{Practical inference throughput on LLaMA2-7B at 50\% retention. All methods use the same device and software stack; reported values aggregate 10 timed repetitions after one warm-up run.}
\label{tab:actual_speedup}
\end{table}

All pruned models achieve over 50\% throughput improvement over the dense baseline, and GPrune-LLM achieves comparable practical speedup to the matched baselines across scoring metrics.

\FloatBarrier

\raggedbottom
\section{Algorithm and Additional Method Details}
\label{sec:app_algorithm}

\subsection{GPrune-LLM Algorithm}

Algorithm~\ref{alg:gprune} presents the GPrune-LLM pruning procedure through BCM, RASA, and MSL. Section~\ref{sec:app_method_details} specifies the corresponding decision rules.

\begin{breakablealgorithm}
\caption{GPrune-LLM Pruning Procedure}
\label{alg:gprune}
\small
\begin{algorithmic}[1]
\Require Pretrained model \(M\); primary and auxiliary calibration datasets \(A\) and \(B\); neuron and attention head scores computed on \(A\) and \(B\) by a base importance metric; target retention \(R^\star\); candidate module counts \(\mathcal{K}\); quantile levels \(\gamma_{\mathrm{drift}}\) and \(\gamma_{\mathrm{score}}\)
\Ensure Pruned model \(M_{\text{pruned}}\)
\Statex \textbf{Component 1: Behavior-Consistent Modularization (BCM)}
\State Compute cross-distribution rank drift magnitude \(d_i\) from neuron scores on \(A\) and \(B\)
\State Set the drift profile \(\boldsymbol{\phi}_i=[d_i]\)
\For{each FFN layer}
    \State Cluster neurons by parameter similarity using KMeans, with the initial module count selected from \(\mathcal{K}\) by silhouette score
\EndFor
\State Measure each module's rank drift variation \(\sigma_k\) across all FFN layers and select the 20\% with the largest variation
\For{each FFN layer}
    \State Split each selected module into two by KMeans on \(d_i\)
\EndFor
\State Use the metric-specific drift profile selected by the direction-admission criterion, augmenting \(\boldsymbol{\phi}_i=[d_i]\) with \(d_i^{\pm}\) only when \(\Delta_{\mathrm{imp}}>\Delta_{\mathrm{disp}}\)
\For{each FFN layer}
    \State Refine the soft memberships \(p_{ik}\) of neurons to modules by minimizing \(\mathcal{L}_{\mathrm{BCM}}\), jointly preserving parameter similarity and drift profile consistency
    \State Obtain hard modules \(\{\mathcal{C}_k\}_{k=1}^{K}\) from \(p_{ik}\)
\EndFor
\Statex \textbf{Component 2: Reliability-Aware Score Adaptation (RASA)}
\State Let \(\mathcal{E}_{\mathrm{FFN}}\) index all resulting modules across FFN layers
\State Characterize each \(k\in\mathcal{E}_{\mathrm{FFN}}\) by its mean local rank drift \(\bar d^{\mathrm{m}}_k\) and mean primary score \(\bar s^{(A)}_k\)
\State Set \(\delta^{\mathrm{drift}}=Q_{\gamma_{\mathrm{drift}}}(\{\bar d^{\mathrm{m}}_k:k\in\mathcal{E}_{\mathrm{FFN}}\})\)
\State Set \(\delta^{\mathrm{score}}=Q_{\gamma_{\mathrm{score}}}(\{\bar s^{(A)}_k:k\in\mathcal{E}_{\mathrm{FFN}}\})\)
\State Select modules that are distribution sensitive and weakly supported, satisfying \(\bar d^{\mathrm{m}}_k\geq\delta^{\mathrm{drift}}\) and \(\bar s^{(A)}_k\leq\delta^{\mathrm{score}}\), as \(\mathcal{K}_{\mathrm{adapt}}\)
\For{each \(k\in\mathcal{K}_{\mathrm{adapt}}\)}
    \State Estimate local ranking repeatability discrepancy \(u_k^{(D)}\) from disjoint subsets of \(D\in\{A,B\}\)
    \State Select the more repeatable scoring source, \(D_k^\star=\arg\min_{D\in\{A,B\}}u_k^{(D)}\)
\EndFor
\State Use scores from \(D_k^\star\) for adapted modules and from \(A\) otherwise
\Needspace{8\baselineskip}
\Statex \textbf{Component 3: Modular Sparsity Learning (MSL)}
\State Assign a learnable threshold \(\tau_k\) to each FFN module and \(\tau^{\mathrm{attn}}\) to each attention layer
\State For attention layers, construct the cross-entropy (CE) loss using both \(A\) and \(B\)
\State For FFN modules, construct adaptive CE using \(A\), and include \(B\) only when its threshold update aligns with that from \(A\) across calibration subsets
\State Combine these CE losses with knowledge distillation (KD) on both \(A\) and \(B\) to form \(\mathcal{L}_{\mathrm{perf}}\)
\State Enforce the global retention target \(R^\star\) with the augmented Lagrangian term \(\mathcal{L}_{\mathrm{constr}}\)
\State Minimize \(\mathcal{L}_{\mathrm{prune}}=\mathcal{L}_{\mathrm{perf}}+\mathcal{L}_{\mathrm{constr}}\) and apply the learned thresholds \(\tau_k\) and \(\tau^{\mathrm{attn}}\) to obtain \(M_{\text{pruned}}\)
\end{algorithmic}
\end{breakablealgorithm}
\raggedbottom

\subsection{Additional Method Details}
\label{sec:app_method_details}

This section specifies the definitions and decision rules omitted from the main text and Algorithm~\ref{alg:gprune}. We first complete the module construction rules in BCM and then detail the supervision used by MSL.

\paragraph{BCM: Module Initialization.}
As described in the main text, BCM initializes modules in two steps. The first clusters neurons by parameter similarity, while the second pre-separates distribution-robust from distribution-sensitive neurons by splitting coarse modules with high internal rank drift variation. We specify the second step here. Let $\mathcal{E}$ index the coarse modules $\mathcal{C}_k$ across all FFN layers. We define their rank drift variation as
\begin{equation}
\sigma_k
=\operatorname{Std}(\{d_i:i\in\mathcal{C}_k\}).
\end{equation}
BCM selects the 20\% of modules with the largest $\sigma_k$ across $\mathcal{E}$ and splits each selected module by two-cluster KMeans on $d_i$.

\paragraph{BCM: Soft Assignment Refinement.}
Following module initialization, BCM refines soft assignments to jointly preserve parameter similarity and rank drift consistency. Because average compactness can underconstrain a small number of diffuse modules, we regularize the modules with the largest pairwise parameter dispersion:
\begin{equation}
\label{eq:l_pair_bcdm_app}
\begin{aligned}
\ell_k
&=
\frac{
\sum_{i\neq j}p_{ik}p_{jk}
\left(1-\cos(\mathbf{w}_i,\mathbf{w}_j)\right)
}{
\sum_{i\neq j}p_{ik}p_{jk}
},\\
\mathcal{L}_{\text{pair}}
&=\frac{1}{|\mathcal{T}|}\sum_{k\in\mathcal{T}}\ell_k,
\end{aligned}
\end{equation}
where $\mathcal{T}$ contains the highest-dispersion quartile of modules. This targeted term prevents diffuse modules from being hidden by the average compactness objective.

The BCM refinement objective $\mathcal{L}_{\mathrm{BCM}}$ is a weighted combination of $\mathcal{L}_{\mathrm{inner}}$, $\mathcal{L}_{\mathrm{pair}}$, $\mathcal{L}_{\mathrm{rep}}$, and $\mathcal{L}_{\mathrm{consis}}$.

\paragraph{BCM: Direction-Aware Refinement.}
Rank drift magnitude captures instability but not which distribution favors a neuron. Signed drift can separate neurons with opposing distribution preferences, but may also weaken within-module drift coherence. To assess whether direction sharpens the module structure relevant to pruning, we use the midpoint of normalized neuron ranks as a representative aggressive pruning boundary. BCM admits direction only when the improvement in separation across this boundary outweighs the accompanying increase in within-module drift variation.

Let $\Delta r_i=(r_A(i)-r_B(i))/(N-1)$ denote signed rank drift. Because magnitude and signed drift form a shared profile, we match their scales within each layer so that neither coordinate dominates $\mathcal{L}_{\mathrm{consis}}$ solely because of its numerical range:
\begin{equation}
d_i^{\pm}
=\Delta r_i
\frac{\operatorname{Std}(d)}{\operatorname{Std}(\Delta r)}.
\end{equation}
Starting from the same initialized modules, we refine a magnitude-only profile $\boldsymbol{\phi}_i^{\mathrm{mag}}=[d_i]$ and a direction-aware profile $\boldsymbol{\phi}_i^{\mathrm{dir}}=[d_i,d_i^{\pm}]$. For the latter, $\mathcal{L}_{\mathrm{consis}}$ averages within-module variation over both coordinates. Let $v\in\{\mathrm{mag},\mathrm{dir}\}$ index the resulting partition $\{\mathcal{C}_k^{(v)}\}_{k=1}^{K_v}$.

Let $\tilde r_D(i)=r_D(i)/(N-1)$ be the normalized rank on distribution $D$. For each partition, the fraction of neurons below this reference boundary within module $k$ is
\begin{equation}
q_k^{(v)}
=\frac{1}{|\mathcal{C}_k^{(v)}|}
\sum_{i\in\mathcal{C}_k^{(v)}}
\mathbf{1}\!\left[
\frac{\tilde r_A(i)+\tilde r_B(i)}{2}<\frac{1}{2}
\right].
\end{equation}
We then measure boundary mixing $I_v$ and within-module drift variation $S_v$:
\begin{equation}
\begin{aligned}
I_v
&=\frac{1}{N_{\mathrm{tot}}}
\sum_{k=1}^{K_v}2|\mathcal{C}_k^{(v)}|
\min\!\left(q_k^{(v)},1-q_k^{(v)}\right),\\
S_v
&=\frac{1}{N_{\mathrm{tot}}}
\sum_{k=1}^{K_v}|\mathcal{C}_k^{(v)}|
\operatorname{Std}_{i\in\mathcal{C}_k^{(v)}}(d_i),
\end{aligned}
\end{equation}
where $N_{\mathrm{tot}}$ is the total number of FFN neurons. Let $\Delta_{\mathrm{imp}}$ denote the positive relative reduction in boundary mixing from the magnitude-only partition to the direction-aware partition, and let $\Delta_{\mathrm{disp}}$ denote the corresponding positive relative increase in drift variation. BCM selects the direction-aware partition only when $\Delta_{\mathrm{imp}}>\Delta_{\mathrm{disp}}$ and otherwise retains the magnitude-only partition. We evaluate this criterion once per base scoring metric and keep the selected profile fixed across model families and retention ratios. It selects the direction-aware profile for twFLAP and the magnitude-only profile for FLAP and Wanda-sp.

\paragraph{MSL: Adaptive Cross-Distribution Supervision.}
After assigning reliable importance scores, MSL learns module-specific pruning thresholds under a global budget. Its performance objective combines CE with KD from the unpruned model. To improve cross-distribution generalization without introducing conflicting supervision, MSL treats these two signals differently. KD matches the full teacher output distribution and provides smoother targets, making the resulting threshold updates less likely to conflict across calibration distributions. We therefore retain KD on both \(A\) and \(B\) throughout threshold learning.

CE follows observed next-token targets and produces a sharper, more distribution-specific signal. Auxiliary CE can therefore oppose the primary update. The module-specific FFN thresholds make a finer sparsity allocation and are more sensitive to such conflicts, whereas attention uses only one threshold per layer. We therefore adapt only FFN CE and retain CE from both distributions for attention. After normalizing the total CE weight, FFN threshold learning chooses between
\begin{equation}
\mathcal{L}_{\mathrm{CE}}^{A}
=\mathcal{L}_{\mathrm{CE}}^{(A)},
\qquad
\mathcal{L}_{\mathrm{CE}}^{A+B}
=\frac{1}{2}
\left(
\mathcal{L}_{\mathrm{CE}}^{(A)}
+\mathcal{L}_{\mathrm{CE}}^{(B)}
\right).
\end{equation}
To test whether the contribution of auxiliary CE is transferable rather than specific to one subset, we evaluate it against the primary update on a disjoint subset. We split each calibration distribution into two disjoint subsets, \(A',A''\) and \(B',B''\), and let \(\mathbf{g}_{X}\) denote the FFN threshold gradient under CE on \(X\). The gradient changes induced by adding CE from the auxiliary distribution on the two subset pairs are
\begin{equation}
\Delta\mathbf{g}_{B'}=\mathbf{g}_{A'+B'}-\mathbf{g}_{A'},
\qquad
\Delta\mathbf{g}_{B''}=\mathbf{g}_{A''+B''}-\mathbf{g}_{A''}.
\end{equation}
MSL includes auxiliary CE only when
\begin{equation}
\left\langle\Delta\mathbf{g}_{B'},\mathbf{g}_{A''}\right\rangle>0,
\qquad
\left\langle\Delta\mathbf{g}_{B''},\mathbf{g}_{A'}\right\rangle>0.
\end{equation}
If either cross-subset alignment fails, MSL retains \(\mathcal{L}_{\mathrm{CE}}^{A}\).

\section{Detailed Experimental Settings}
\label{sec:detailed_settings}

This section details our experimental configuration. Table~\ref{tab:hyperparams} summarizes the principal hyperparameters, and the remaining settings are described below.

\begin{table}[H]
\centering
{\fontsize{9}{10.4}\selectfont
\setlength{\tabcolsep}{2.5pt}
\begin{tabular}{@{}l p{0.48\columnwidth} p{0.24\columnwidth}@{}}
\toprule
\textbf{Component} & \textbf{Parameter} & \textbf{Value} \\
\midrule
\multirow{4}{*}{General} & Calibration samples for each of $A$ and $B$ & 2048 \\
 & Calibration sequence length & 128 \\
 & Primary calibration dataset $A$ & WikiText2 \\
 & Auxiliary calibration dataset $B$ & PTB \\
\midrule
BCM & Candidate initial module counts $\mathcal{K}$ & 16, 24, 32, 40, 48 \\
\midrule
\multirow{2}{*}{RASA} & Rank-drift quantile level $\gamma_{\mathrm{drift}}$ & 0.90 \\
 & Primary-score quantile level $\gamma_{\mathrm{score}}$ & 0.90 (0.85 for Vicuna1.5-7B) \\
\bottomrule
\end{tabular}
}
\caption{Principal hyperparameters used in the reported experiments.}
\label{tab:hyperparams}
\end{table}

\noindent\textbf{BCM.}
Most scoring metrics assign neuron importance independently of the module partition, so BCM can use their rankings directly. twFLAP instead couples scoring with the partition because its token-aware scores are aggregated within modules. We therefore use FLAP rankings to form an initial partition, refine it with the resulting twFLAP rankings, and recompute twFLAP scores on the final modules.

\noindent\textbf{RASA.}
We explore each RASA quantile from $0.80$ to $0.90$ with a step size of $0.05$. Across models, the rank-drift sweep consistently favors $\gamma_{\mathrm{drift}}=0.90$, so we use this value throughout. The preferred primary-score quantile varies by model, with $\gamma_{\mathrm{score}}=0.90$ for LLaMA2 and $0.85$ for Vicuna1.5. Once selected for a model, both quantile levels are fixed across base scoring metrics and retention ratios. The model-dependent preference for $\gamma_{\mathrm{score}}$ may be related to the alignment between each model's training distribution and the primary calibration distribution. Such alignment may shift the score level at which primary support is sufficiently reliable to retain a module's original ranking.

\noindent\textbf{MSL.}
We use a fixed convergence-based schedule of 2--3 threshold learning epochs, with one additional epoch for LLaMA2-13B to accommodate the slower convergence observed for the larger model.

\noindent\textbf{Baselines.}
We evaluate FANG with its functional grouping, token-weighted FLAP scoring, and sparsity allocation using the primary calibration distribution. We disable OBC reconstruction to ensure a fair comparison under the same one-shot pruning protocol without weight reconstruction. GPrune-LLM (twFLAP) uses the same scoring family within behavior-consistent modules and completes pruning with RASA and MSL. The comparison therefore evaluates whether cross-distribution restructuring improves generalization beyond token-aware scoring.

\noindent\textbf{Evaluation.}
In the EleutherAI LM Evaluation Harness benchmark, PIQA, HellaSwag, ARC-e, ARC-c, and OBQA provide both raw and length-normalized accuracy. For these tasks, we report the higher value for every method. Due to computational resource constraints, we report one run for each pruning configuration. Unless otherwise specified in the multi-seed study, all runs use the fixed random seed 0.

\section{Robustness to the Evaluation Protocol}
\label{sec:accuracy_protocol_robustness}

As specified in our evaluation setup (Appendix~\ref{sec:detailed_settings}), our main results take the higher of raw accuracy (\texttt{acc}) and length-normalized accuracy (\texttt{acc\_norm}) for each task. Although this rule is applied uniformly, which of the two is higher can differ across methods and settings, so the actual evaluation metric used for a given task may not be the same in every comparison. To avoid any incidental overestimation this rule may introduce, we recompute all task results under a fully deterministic protocol that fixes the evaluation metric per task, using \texttt{acc\_norm} when available and \texttt{acc} otherwise. Tables~\ref{tab:llama2_7b_fang_metric}--\ref{tab:llama2_13b_fang_metric} reuse the same checkpoints and evaluation outputs as the main results, so only the reported task metrics and their averages change.

\begingroup
\renewcommand{\arraystretch}{0.80}

\begin{table}[H]
\centering
{\fontsize{9}{10.4}\selectfont
\setlength{\tabcolsep}{1.8pt}
\begin{tabular}{@{}c l | c c | c c c c c c c | c@{}}
\hline
Ratio & Method & WT2 PPL$\downarrow$ & PTB PPL$\downarrow$ & BoolQ & PIQA & HellaSwag & WinoGrande & ARC-e & ARC-c & OBQA & Avg. Acc.$\uparrow$ \\
\hline
100\% & LLaMA2-7B & 12.19 & 48.35 & 71.10 & 76.82 & 72.96 & 67.25 & 53.58 & 40.53 & 40.80 & 60.43 \\
\hline
80\% & LLM-Pruner & 19.23 & 72.62 & 62.26 & \textbf{75.41} & 65.82 & 60.85 & 50.17 & 37.46 & 39.20 & 55.88 \\
 & Wanda-sp & 17.50 & 67.01 & 46.79 & \underline{75.14} & \underline{68.13} & 65.43 & \underline{51.64} & \textbf{38.82} & 38.20 & 54.88 \\
 & GPrune-LLM (Wanda-sp) & 15.44$\uparrow$ & \textbf{56.95}$\uparrow$ & 60.76$\uparrow$ & 74.54 & \textbf{68.67}$\uparrow$ & \underline{65.51}$\uparrow$ & 50.55 & 37.80 & \underline{39.80}$\uparrow$ & 56.80$\uparrow$ \\
 & FLAP & 16.41 & 61.61 & 52.45 & 73.01 & 62.56 & 62.12 & 47.35 & 34.81 & 39.20 & 53.07 \\
 & GPrune-LLM (FLAP) & 15.55$\uparrow$ & 59.56$\uparrow$ & 64.16$\uparrow$ & 74.76$\uparrow$ & 66.49$\uparrow$ & 62.19$\uparrow$ & 49.37$\uparrow$ & 36.35$\uparrow$ & 39.40$\uparrow$ & 56.10$\uparrow$ \\
 & FANG & \underline{15.43} & 61.55 & \underline{67.16} & 74.32 & 67.34 & \textbf{65.67} & 50.80 & 38.14 & \textbf{40.20} & \underline{57.66} \\
 & GPrune-LLM (twFLAP) & \textbf{15.33}$\uparrow$ & \underline{58.19}$\uparrow$ & \textbf{67.89}$\uparrow$ & 74.16 & 67.90$\uparrow$ & 65.43 & \textbf{53.28}$\uparrow$ & \underline{38.23}$\uparrow$ & 39.40 & \textbf{58.04}$\uparrow$ \\
\hline
70\% & LLM-Pruner & 29.40 & 111.56 & 50.15 & 71.82 & 55.64 & 55.49 & 44.23 & 31.57 & 36.40 & 49.33 \\
 & Wanda-sp & 22.94 & 79.79 & 58.78 & 70.29 & 59.44 & 55.56 & 48.23 & \textbf{37.03} & \underline{39.40} & 52.68 \\
 & GPrune-LLM (Wanda-sp) & 21.04$\uparrow$ & 75.56$\uparrow$ & 54.80 & 71.60$\uparrow$ & \textbf{62.23}$\uparrow$ & 57.22$\uparrow$ & \underline{48.74}$\uparrow$ & 35.67 & 38.20 & 52.64 \\
 & FLAP & 20.25 & 76.02 & 53.39 & 69.48 & 55.49 & 60.30 & 44.19 & 32.59 & 38.20 & 50.52 \\
 & GPrune-LLM (FLAP) & 18.97$\uparrow$ & \underline{69.94}$\uparrow$ & \underline{63.73}$\uparrow$ & 72.03$\uparrow$ & 60.06$\uparrow$ & 57.30 & 47.98$\uparrow$ & 34.98$\uparrow$ & 38.20 & 53.47$\uparrow$ \\
 & FANG & \underline{18.86} & 78.23 & 61.65 & \textbf{72.14} & \underline{61.68} & \textbf{62.12} & \textbf{50.17} & \underline{36.77} & 38.60 & \underline{54.73} \\
 & GPrune-LLM (twFLAP) & \textbf{18.76}$\uparrow$ & \textbf{66.84}$\uparrow$ & \textbf{65.02}$\uparrow$ & \underline{72.14} & 61.49 & \underline{61.25} & 48.70 & 36.26 & \textbf{39.80}$\uparrow$ & \textbf{54.95}$\uparrow$ \\
\hline
50\% & LLM-Pruner & 227.85 & 459.50 & \underline{60.61} & 57.78 & 30.56 & 49.33 & 30.35 & 27.05 & 31.40 & 41.01 \\
 & Wanda-sp & 262.32 & 505.51 & 59.82 & 52.39 & 26.54 & 49.41 & 29.00 & 23.89 & 32.80 & 39.12 \\
 & GPrune-LLM (Wanda-sp) & 60.80$\uparrow$ & 191.99$\uparrow$ & 52.72 & 61.75$\uparrow$ & 42.30$\uparrow$ & 52.49$\uparrow$ & 36.57$\uparrow$ & 28.07$\uparrow$ & \textbf{35.80}$\uparrow$ & 44.24$\uparrow$ \\
 & FLAP & 42.81 & 152.49 & 39.63 & 62.13 & 39.54 & 50.36 & 34.09 & 28.24 & 35.20 & 41.31 \\
 & GPrune-LLM (FLAP) & 37.02$\uparrow$ & \textbf{125.41}$\uparrow$ & 53.27$\uparrow$ & \underline{62.57}$\uparrow$ & \textbf{44.20}$\uparrow$ & \textbf{54.78}$\uparrow$ & \textbf{38.80}$\uparrow$ & \underline{29.95}$\uparrow$ & 35.20 & \underline{45.54}$\uparrow$ \\
 & FANG & \textbf{36.31} & 154.23 & \textbf{61.59} & 62.02 & 39.48 & 52.41 & 38.38 & 27.65 & 35.00 & 45.22 \\
 & GPrune-LLM (twFLAP) & \underline{36.83} & \underline{129.77}$\uparrow$ & 57.34 & \textbf{65.23}$\uparrow$ & \underline{43.72}$\uparrow$ & \underline{52.72}$\uparrow$ & \underline{38.76}$\uparrow$ & \textbf{31.74}$\uparrow$ & \underline{35.80}$\uparrow$ & \textbf{46.47}$\uparrow$ \\
\hline
\end{tabular}
}
\caption{Full comparison on LLaMA2-7B under the fixed task-metric reporting protocol.}
\label{tab:llama2_7b_fang_metric}
\end{table}

\renewcommand{\arraystretch}{0.88}
\begin{table}[H]
\centering
{\fontsize{9}{10.4}\selectfont
\setlength{\tabcolsep}{1.8pt}
\begin{tabular}{@{}c l | c c | c c c c c c c | c@{}}
\hline
Ratio & Method & WT2 PPL$\downarrow$ & PTB PPL$\downarrow$ & BoolQ & PIQA & HellaSwag & WinoGrande & ARC-e & ARC-c & OBQA & Avg. Acc.$\uparrow$ \\
\hline
100\% & Vicuna1.5-7B & 16.24 & 60.78 & 75.75 & 77.80 & 71.02 & 67.64 & 56.06 & 39.93 & 42.40 & 61.52 \\
\hline
80\% & LLM-Pruner & 24.90 & 86.20 & 50.98 & \textbf{75.24} & 64.14 & 59.83 & 53.07 & 37.12 & 40.00 & 54.34 \\
 & Wanda-sp & 23.66 & 73.67 & \textbf{67.31} & 73.45 & 65.21 & 61.80 & 53.11 & \underline{38.31} & 39.80 & 57.00 \\
 & GPrune-LLM (Wanda-sp) & 20.12$\uparrow$ & \textbf{67.35}$\uparrow$ & 56.76 & 74.59$\uparrow$ & \underline{66.85}$\uparrow$ & 65.11$\uparrow$ & 53.07 & 36.35 & 39.20 & 55.99 \\
 & FLAP & 21.06 & 76.42 & \underline{66.88} & 71.71 & 63.17 & 64.48 & 51.47 & 34.73 & 39.80 & 56.04 \\
 & GPrune-LLM (FLAP) & \underline{18.95}$\uparrow$ & \underline{68.24}$\uparrow$ & 59.11 & 74.54$\uparrow$ & \textbf{67.26}$\uparrow$ & 65.27$\uparrow$ & 52.57$\uparrow$ & 35.41$\uparrow$ & \textbf{41.60}$\uparrow$ & 56.54$\uparrow$ \\
 & FANG & 19.30 & 74.58 & 60.31 & \underline{74.97} & 66.17 & \underline{65.43} & \textbf{54.04} & \textbf{38.57} & \underline{40.40} & \underline{57.13} \\
 & GPrune-LLM (twFLAP) & \textbf{18.71}$\uparrow$ & 69.15$\uparrow$ & 66.33$\uparrow$ & 73.23 & 66.22$\uparrow$ & \textbf{65.98}$\uparrow$ & \underline{53.87} & 36.60 & 39.80 & \textbf{57.43}$\uparrow$ \\
\hline
70\% & LLM-Pruner & 36.44 & 120.82 & 43.94 & \textbf{72.69} & 56.13 & 57.30 & 48.02 & 34.30 & 38.20 & 50.08 \\
 & Wanda-sp & 32.49 & 98.83 & 44.01 & 71.33 & 57.85 & 56.51 & 50.93 & \underline{37.12} & 38.80 & 50.93 \\
 & GPrune-LLM (Wanda-sp) & 25.34$\uparrow$ & 86.05$\uparrow$ & 51.71$\uparrow$ & 71.65$\uparrow$ & \textbf{62.64}$\uparrow$ & 61.09$\uparrow$ & 49.96 & 36.09 & 39.40$\uparrow$ & 53.22$\uparrow$ \\
 & FLAP & 25.26 & 90.87 & \textbf{66.18} & 69.86 & 57.56 & \underline{62.35} & 48.15 & 33.11 & 38.40 & 53.66 \\
 & GPrune-LLM (FLAP) & \textbf{22.38}$\uparrow$ & \textbf{77.41}$\uparrow$ & 56.39 & 71.55$\uparrow$ & 61.83$\uparrow$ & 61.80 & \underline{51.30}$\uparrow$ & 36.09$\uparrow$ & \textbf{40.20}$\uparrow$ & 54.17$\uparrow$ \\
 & FANG & \underline{22.55} & 91.04 & 53.85 & \underline{71.87} & 61.83 & \textbf{63.69} & \textbf{53.32} & \textbf{37.12} & \underline{40.00} & \underline{54.53} \\
 & GPrune-LLM (twFLAP) & 22.78 & \underline{82.90}$\uparrow$ & \underline{63.24}$\uparrow$ & 70.95 & \underline{62.07}$\uparrow$ & 61.17 & 51.01 & 36.60 & 39.40 & \textbf{54.92}$\uparrow$ \\
\hline
50\% & LLM-Pruner & 168.91 & 450.48 & \textbf{62.39} & 60.23 & 34.12 & 53.35 & 32.87 & 27.82 & 34.00 & 43.54 \\
 & Wanda-sp & 270.82 & 551.68 & 60.83 & 54.30 & 27.38 & 50.12 & 29.00 & 23.04 & 31.20 & 39.41 \\
 & GPrune-LLM (Wanda-sp) & 71.07$\uparrow$ & 304.31$\uparrow$ & 44.04 & 62.95$\uparrow$ & 42.49$\uparrow$ & 52.33$\uparrow$ & 41.50$\uparrow$ & \textbf{31.66}$\uparrow$ & 33.80$\uparrow$ & 44.11$\uparrow$ \\
 & FLAP & 50.23 & 179.31 & 38.50 & \underline{63.22} & 41.22 & 53.28 & 41.33 & 29.35 & \underline{34.60} & 43.07 \\
 & GPrune-LLM (FLAP) & \textbf{43.62}$\uparrow$ & \textbf{123.00}$\uparrow$ & 58.87$\uparrow$ & \textbf{64.47}$\uparrow$ & \textbf{44.02}$\uparrow$ & 52.80 & \textbf{43.39}$\uparrow$ & \underline{30.80}$\uparrow$ & \textbf{37.20}$\uparrow$ & \textbf{47.36}$\uparrow$ \\
 & FANG & \underline{45.31} & 183.42 & 46.94 & 62.73 & 42.23 & \underline{54.54} & 41.79 & 28.33 & 34.60 & 44.45 \\
 & GPrune-LLM (twFLAP) & 51.81 & \underline{143.59}$\uparrow$ & \underline{62.08}$\uparrow$ & 62.79$\uparrow$ & \underline{43.23}$\uparrow$ & \textbf{54.62}$\uparrow$ & \underline{42.34}$\uparrow$ & 29.18$\uparrow$ & 34.40 & \underline{46.95}$\uparrow$ \\
\hline
\end{tabular}
}
\caption{Full comparison on Vicuna1.5-7B under the fixed task-metric reporting protocol.}
\label{tab:vicuna15_7b_fang_metric}
\end{table}

\begin{table}[H]
\centering
{\fontsize{9}{10.4}\selectfont
\setlength{\tabcolsep}{1.8pt}
\begin{tabular}{@{}c l | c c | c c c c c c c | c@{}}
\hline
Ratio & Method & WT2 PPL$\downarrow$ & PTB PPL$\downarrow$ & BoolQ & PIQA & HellaSwag & WinoGrande & ARC-e & ARC-c & OBQA & Avg. Acc.$\uparrow$ \\
\hline
100\% & LLaMA2-13B & 10.98 & 54.42 & 68.99 & 79.05 & 76.60 & 69.77 & 57.95 & 44.20 & 42.00 & 62.65 \\
\hline
50\% & Wanda-sp & 205.58 & 393.96 & 61.16 & 50.87 & 26.65 & 49.49 & 28.66 & 26.19 & 29.60 & 38.95 \\
 & GPrune-LLM (Wanda-sp) & 28.30$\uparrow$ & 126.67$\uparrow$ & 62.17$\uparrow$ & \textbf{67.90}$\uparrow$ & \textbf{54.20}$\uparrow$ & 55.09$\uparrow$ & 43.39$\uparrow$ & \textbf{33.62}$\uparrow$ & \underline{39.00}$\uparrow$ & \underline{50.77}$\uparrow$ \\
 & FLAP & 27.91 & 156.91 & \textbf{62.66} & 65.02 & 48.95 & \textbf{57.85} & 43.22 & 31.48 & 37.00 & 49.46 \\
 & GPrune-LLM (FLAP) & 26.54$\uparrow$ & \textbf{105.20}$\uparrow$ & 61.59 & 67.52$\uparrow$ & \underline{53.71}$\uparrow$ & 56.75 & \underline{44.32}$\uparrow$ & 32.85$\uparrow$ & 38.20$\uparrow$ & 50.71$\uparrow$ \\
 & FANG & \underline{26.43} & 155.46 & \underline{62.35} & 64.85 & 50.53 & 55.64 & 43.60 & 31.23 & \textbf{39.20} & 49.63 \\
 & GPrune-LLM (twFLAP) & \textbf{25.00}$\uparrow$ & \underline{120.01}$\uparrow$ & 62.26 & \underline{67.85}$\uparrow$ & 51.22$\uparrow$ & \underline{56.99}$\uparrow$ & \textbf{45.58}$\uparrow$ & \underline{33.19}$\uparrow$ & 38.40 & \textbf{50.78}$\uparrow$ \\
\hline
40\% & Wanda-sp & 249.44 & 565.56 & 61.68 & 50.54 & 26.42 & 49.64 & 27.44 & 24.91 & 30.60 & 38.75 \\
 & GPrune-LLM (Wanda-sp) & 83.47$\uparrow$ & 357.04$\uparrow$ & 61.74$\uparrow$ & 60.61$\uparrow$ & 40.44$\uparrow$ & 51.54$\uparrow$ & \underline{37.37}$\uparrow$ & 26.62$\uparrow$ & \underline{35.80}$\uparrow$ & \underline{44.88}$\uparrow$ \\
 & FLAP & 45.29 & 263.68 & \underline{62.08} & 58.11 & 36.90 & 52.25 & 35.56 & 27.39 & 34.80 & 43.87 \\
 & GPrune-LLM (FLAP) & \underline{45.21}$\uparrow$ & \textbf{154.85}$\uparrow$ & 49.97 & \underline{60.99}$\uparrow$ & \underline{40.83}$\uparrow$ & 52.01 & 36.11$\uparrow$ & \underline{29.61}$\uparrow$ & 35.40$\uparrow$ & 43.56 \\
 & FANG & \textbf{39.51} & 230.79 & 57.98 & 58.92 & 39.89 & \underline{52.49} & 36.57 & 27.73 & 35.00 & 44.08 \\
 & GPrune-LLM (twFLAP) & 46.04 & \underline{172.63}$\uparrow$ & \textbf{62.11}$\uparrow$ & \textbf{61.26}$\uparrow$ & \textbf{41.27}$\uparrow$ & \textbf{53.67}$\uparrow$ & \textbf{40.91}$\uparrow$ & \textbf{30.63}$\uparrow$ & \textbf{36.80}$\uparrow$ & \textbf{46.66}$\uparrow$ \\
\hline
30\% & Wanda-sp & 1943.76 & 3163.66 & 51.16 & 49.67 & 25.54 & 47.51 & 26.35 & \underline{27.39} & 29.20 & 36.69 \\
 & GPrune-LLM (Wanda-sp) & 362.63$\uparrow$ & 1023.02$\uparrow$ & 52.45$\uparrow$ & 54.68$\uparrow$ & \underline{31.20}$\uparrow$ & \underline{52.01}$\uparrow$ & 32.70$\uparrow$ & 26.88 & \textbf{34.80}$\uparrow$ & \underline{40.67}$\uparrow$ \\
 & FLAP & 2676.82 & 3872.98 & 37.74 & 50.05 & 25.61 & 48.78 & 27.48 & \textbf{28.92} & 29.60 & 35.46 \\
 & GPrune-LLM (FLAP) & 119.35$\uparrow$ & \textbf{339.25}$\uparrow$ & 47.68$\uparrow$ & \underline{54.73}$\uparrow$ & 30.80$\uparrow$ & \textbf{53.43}$\uparrow$ & \textbf{33.59}$\uparrow$ & 27.22 & 32.60$\uparrow$ & 40.01$\uparrow$ \\
 & FANG & \textbf{75.39} & 392.39 & \underline{54.53} & 54.30 & 30.72 & 50.36 & 31.02 & 25.60 & 32.00 & 39.79 \\
 & GPrune-LLM (twFLAP) & \underline{90.77} & \underline{391.91}$\uparrow$ & \textbf{59.51}$\uparrow$ & \textbf{55.17}$\uparrow$ & \textbf{31.58}$\uparrow$ & 51.85$\uparrow$ & \underline{32.91}$\uparrow$ & 26.62$\uparrow$ & \underline{34.20}$\uparrow$ & \textbf{41.69}$\uparrow$ \\
\hline
\end{tabular}
}
\caption{Full comparison on LLaMA2-13B under the fixed task-metric reporting protocol.}
\label{tab:llama2_13b_fang_metric}
\end{table}
\endgroup

The comparative conclusions still hold under the fixed protocol. GPrune-LLM remains the best-performing method in all nine model--retention settings and still improves its corresponding base metric in the large majority of them. The improvements remain larger at higher sparsity, and the accuracy gap across base metrics still shrinks after applying GPrune-LLM.

\end{document}